\documentclass[10pt,twocolumn,letterpaper]{article}

\usepackage{cvpr}              

\definecolor{cvprblue}{rgb}{0.21,0.49,0.74}
\usepackage[pagebackref,breaklinks,colorlinks,allcolors=cvprblue]{hyperref}

\usepackage{url,lineno,microtype,subcaption}
\usepackage{multirow}
\usepackage{xspace}
\usepackage{appendix}
\usepackage{makecell}
\usepackage{ragged2e}
\usepackage{float}
\newcommand{\Video}[1]{V_{#1}}
\newcommand{\RWindow}[1]{\mathcal{R}_{#1}}
\newcommand{\QueryFrame}[1]{\mathcal{Q}_{#1}}
\newcommand{\QueryMask}[1]{\mathcal{M}_{#1}}
\newcommand{\SynFrames}{\ensuremath{\mathbb{F}}\relax}
\newcommand{\WheatVideos}{\ensuremath{\mathbb{W}}\relax}
\newcommand{\PseudoLabeledVideos}[1]{\ensuremath{\mathbb{P}}_{#1}\relax}
\newcommand{\ValidSingleDomainSamples}{\ensuremath{\Delta}\relax}
\newcommand{\TestSingleDomain}{\ensuremath{\Psi}\relax}
\newcommand{\TestDroneSamples}{\ensuremath{\Gamma}\relax}

\newcommand{\ImageSyntheticModel}[1]{\ensuremath{\mathbb{IM}_{\text{Synt}}^{#1}}{}\xspace}
\newcommand{\ImagePseudoModel}[1]{\ensuremath{\mathbb{IM}_{\text{Pseu}}^{#1}}{}\xspace}
\newcommand{\VideoSyntheticModel}[1]{\ensuremath{\mathbb{VM}_{\text{Synt}}^{#1}}{}\xspace}
\newcommand{\VideoPseudoModel}[1]{\ensuremath{\mathbb{VM}_{\text{Pseu}}^{#1}}{}\xspace}
\newcommand{\XMemSyntheticModel}{\ensuremath{\mathbb{XM}_{\text{Synt}}}{}\xspace}
\newcommand{\XMemPseudoModel}{\ensuremath{\mathbb{XM}_{\text{Pseu}}}{}\xspace}

\def\paperID{15} 
\def\confName{Vision for Agriculture}
\def\confYear{2025}

\title{A Semi-Self-Supervised Approach for Dense-Pattern Video Object Segmentation}

\author{Keyhan Najafian$^1$, Farhad Maleki$^2$, Lingling Jin$^1$, Ian Stavness$^1$\\
$^1$Department of Computer Science, University of Saskatchewan, Saskatoon, Saskatchewan, Canada \\
$^2$Department of Computer Science, University of Calgary, Calgary, Alberta, Canada \\
{\tt\small \{keyhan.najafian, lingling.jin, ian.stavness\}@usask.ca, farhad.maleki1@ucalgary.ca}
}

\begin{document}
\maketitle
\begin{abstract}
Video object segmentation (VOS)---predicting pixel-level regions for objects within each frame of a video---is particularly challenging in agricultural scenarios, where videos of crops include hundreds of small, dense, and occluded objects (stems, leaves, flowers, pods) that sway and move unpredictably in the wind. Supervised training is the state-of-the-art for VOS, but it requires large, pixel-accurate, human-annotated videos, which are costly to produce for videos with many densely packed objects in each frame. To address these challenges, we proposed a semi-self-supervised spatiotemporal approach for \textit{dense}-VOS (DVOS) using a diffusion-based method through multi-task (reconstruction and segmentation) learning. We train the model first with synthetic data that mimics the camera and object motion of real videos and then with pseudo-labeled videos. We evaluate our DVOS method for wheat head segmentation from a diverse set of videos (handheld, drone-captured, different field locations, and different growth stages---spanning from Boot-stage to Wheat-mature and Harvest-ready). Despite using only a few manually annotated video frames, the proposed approach yielded a high-performing model, achieving a Dice score of $0.79$ when tested on a drone-captured external test set. While our method was evaluated on wheat head segmentation, it can be extended to other crops and domains, such as crowd analysis or microscopic image analysis.

\end{abstract}    
\section{Introduction}\label{sec:intro}
Video object segmentation (VOS) is a fundamental computer vision task that involves automatically extracting and precisely delineating objects of interest at the pixel level across consecutive frames in a video~\cite{cheng2022xmem}. In VOS, the association between pixels and objects can evolve over time due to object motion, camera movement, or change in perspective. In contrast to static images, videos capture the dynamic nature of motion and interaction as they unfold temporally. However, videos often contain significant noise and variability, influenced by uncontrollable environmental factors, such as camera motion, sensor limitations, wind, and challenging lighting conditions (i.e.\ harsh sunlight). These issues are particularly pronounced in agricultural contexts, where videos frequently exhibit these characteristics, posing challenges for DL-based analysis.\par

VOS is often more effective than static image analysis. For instance, single-frame analysis can introduce artifacts and struggle to address challenges like background clutter and transient objects~\cite{koprinska2001temporal, NguyenThai2021ASA}. This is particularly evident in agricultural settings, where shadows from swaying wheat heads can degrade model performance.
%
%
%
Also, unlike general video tasks, where objects are larger and move predictably~\cite{cheng2022xmem},  agricultural scenes often feature many small, self-similar objects (e.g.\ wheat heads) that shift erratically across frames~\cite{Najafian2022SemiSelfSupervisedLF}. This unpredictability complicates optical flow-based methods and makes pixel-level classification particularly difficult, posing significant hurdles for accurate segmentation in such dynamic environments. \par

Developing high-performing DL models for VOS is complex, requiring solutions to address limitations such as data availability, annotation costs, model architecture design, and computational resources~\cite{ho2022imagen, tan2023temporal}. Supervised learning, which relies on large-scale annotated datasets, is resource-intensive, especially for pixel-level tasks including VOS. Manual annotation of videos, which can contain thousands of frames, is laborious and expensive, especially in agricultural scenes, where images often feature dense and repetitive patterns of objects such as leaves, stems, flowers, and wheat spikes ~\cite{lipton2021zoetrope, real2017youtube,Najafian2022SemiSelfSupervisedLF}.\par

To address these challenges, recent research has explored semi-supervised~\cite{yang2022survey} and self-supervised~\cite{ciga2022self,ho2020denoising} methods for tasks such as wheat head detection~\cite{Fourati2021WheatHD} and segmentation~\cite{myers2024modified}. 
Building on prior advancements in semi- and self-supervised learning for wheat head analysis, we propose a novel semi-self-supervised training methodology for DVOS. This approach focuses on wheat head segmentation in videos characterized by dense, repetitive patterns across various growth stages and only requires a few human-annotated video frames. \par

%
Our approach synthesizes a large-scale dataset of computationally annotated videos, eliminating the need for extensive manual annotation. We propose a UNet-style~\cite{ronneberger2015u} architecture for multi-task training, with a two-stage process: first, construct a foundational model using synthesized videos (that mimic camera and objection motion), and then fine-tune it with pseudo-labeled videos, where the labels generated frame by frame by an image-based model~\cite{Najafian2022SemiSelfSupervisedLF}. This methodology ensures robust performance while significantly reducing the annotation effort.\par
We evaluate our proposed approach on three test sets: (1) a small-scale, single-domain, pixel-accurate video dataset; (2) a diverse, large-scale wheat field video dataset captured with handheld cameras and semi-automatically labeled, model-generated and human-validated selection; and (3) a drone-captured, manually annotated wheat field video dataset. This demonstrates the utility of our method in addressing the limitations of dense-pattern VOS in agricultural contexts. \par
The key contributions of this work include: (1) the design of a convolution-based architecture enhanced with diffusion and attention mechanisms for DVOS; (2) the development of a data synthesis pipeline for generating large-scale and computationally annotated videos that simulate camera and object motion in natural environments; (3) the creation of a large-scale and diverse dataset of wheat field videos with computationally-generated annotations, utilized for further model training; (4) the formulation of the DVOS task for wheat head video data, enabling predictions without the need for reference frame initialization; and (5) a comprehensive evaluation of the proposed model across phenotypically diverse datasets, demonstrating superior generalization across varying data domains compared to the conventional VOS approaches. The source code for our method is publicly available at~\href{https://github.com/USask-BINFO/DVOS.git}{GitHub}.

\section{Related Work}
General video object segmentation methods have been investigated by analyzing the spatial and temporal characteristics inherent to video data. 
While the state-of-the-art VOS approaches commonly opt to compute temporal matching in the form of optical flow and dense trajectories~\cite{cheng2022xmem,zhang2023boosting}, others prefer the parallel strategy that processes frames independently~\cite{perazzi2017learning,maninis2018video,robinson2020learning}.
These studies rely on the availability of large-scale annotated datasets, i.e., the pixel-level annotation for each target object in every video frame, to serve as the primary input for training DL models~\cite{cheng2022xmem}. \par 
It has been reported that for general-purpose tasks involving one or two large target objects, these models often take shortcuts by predicting subsequent masks solely based on the provided preceding frame's mask, thereby bypassing the rich information offered by the preceding frames themselves~\cite{zhang2023boosting}. This oversight led to shortcomings in predicting segmentation maps, stemming from obscured objects in earlier frames and missed objects within the most recent mask~\cite{maninis2018video}. Recurrent~\cite{Ventura2019RVOSER} and memory-bank-based~\cite{cheng2022xmem} are examples of one-shot VOS methods that use an initial reference mask to predict subsequent masks, updating them as hidden states or storing them in memory banks for future frame predictions. \par
Certain studies suggest leveraging solely temporal signals for object segmentation. For instance, a recent work~\cite{yang2021self} utilized optical flow extracted from reference frames to trace the main object in the query frame, employing a straightforward transformer-based architecture. \par 
Using spatial or temporal information, video data has widespread applications in precision agriculture~\cite{campos2016spatio,Liu2021SpatialSF}.
Campos et al.~\cite{campos2016spatio} employed conventional image processing methods to detect static and dynamic obstacles. Through obstacle segmentation and detection, they analyzed spatiotemporal information extracted from videos captured by cameras mounted on mobile vehicles in agricultural environments. \par
The study of image-based segmentation on images with dense and repetitive patterns~\cite{myers2024modified, Heschl2024SynthSetGD} has been explored within the agricultural domain, ranging from smaller-scale investigations~\cite{das2021deepveg} to larger-scale data analyses, with examples such as wheat head detection~\cite{david2021global}, counting~\cite{ubbens2020autocount}, and segmentation~\cite{tan2020rapid}.
Sabzi et al.~\cite{sabzi2019use} employed traditional image processing techniques---utilizing intensity transformations and morphological operations on image color and texture features---to segment agricultural video frames, individually, characterized by complex and dense patterns. 
Ariza-Sentis et al.~\cite{ariza2023object} utilized phenotyping techniques to assess the physical characteristics, including size, shape, and quantity, of grape bunches and berries. This involved employing multi-object tracking and instance segmentation (spatial embedding) methods to determine the attributes of individual white grape bunches and berries from RGB videos captured by unmanned aerial vehicles flying over a commercial vineyard densely covered with leaves. Gibbs et al.~\cite{Gibbs2019RecoveringWP} aimed to create a generalizable feature detection method combined with a tracking algorithm to enhance feature detection and enable the determination of plant movement traits. 
\section{Method}\label{sec:method}
%
\subsection{Problem Formulation}
\begin{figure}[!tbh]
    \centering
    \includegraphics[width=\linewidth]{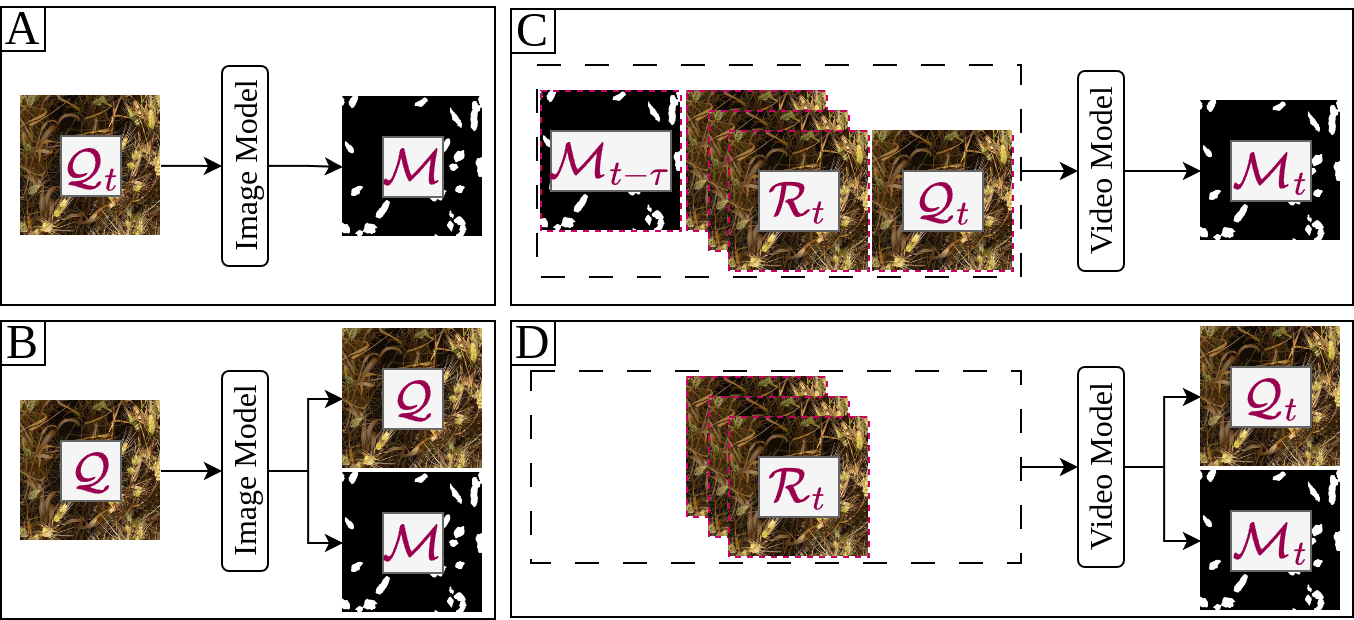}
    \caption{Architectural choices for segmentation: (A) Conventional image segmentation, segmenting the query image~\cite{ronneberger2015u}; (B) Multi-task learning, jointly learning image reconstruction and image segmentation~\cite{ghanbari2024semi}; (C) Conventional VOS, using reference frames, first reference frame mask, and query frame as model inputs to segment the query image~\cite{Wang2021SwiftNetRV}; (D) Our approach uniquely leverages reference frames within a multitasking framework to predict both the subsequent query frame and its corresponding mask, eliminating the need for reference frame annotations or the query frame itself as model input.
    }
    \label{fig:main_figure_01}
\end{figure}

In general, VOS aims to generate a semantic segmentation mask for each frame in a video clip of length $\tau$. Variations of VOS problem settings include image-based models, which predict a segmentation mask for each frame independently (Figure~\ref{fig:main_figure_01}A). Multi-task image-based models~\cite{ghanbari2024semi} attempt to improve model training and performance by forcing the model to simultaneously predict masks and a reconstructed version of the input image (Figure~\ref{fig:main_figure_01}B). The frame-by-frame approaches, however, are identical to the single image problem setting and do not incorporate information across frames to assist in the predictions. In the traditional VOS setting, the model predicts the segmentation mask for the query (subsequent) frame based on the preceding frames, their first frame's mask, and the query frame itself as input (Figure~\ref{fig:main_figure_01}C). Our proposed multi-task approach, however, generates the query frame and predicts its segmentation mask given only the preceding frames as input, Figure~\ref{fig:main_figure_01}D.

We define the following terminology to describe the frames within a video clip.
Given a sequence of frames $\{x_r\mid t- \tau\leq r \leq t\}$, we refer to $\RWindow{t}=\{x_r\mid t- \tau\leq r \leq t-1\}$ as reference window, where $\tau$ is the context window representing temporal reference information. Denote the query frame of $x_t$ as $\QueryFrame{t}$ and the pixel-level annotation for the query frame $\QueryFrame{t}$ as $\QueryMask{t}$.
$\RWindow{t}$ is used as input for the model, and the pair $(\QueryFrame{t}, \QueryMask{t})$ is the desired output, i.e., the ground truth (Figure~\ref{fig:main_figure_01}D). \par
%
We adopt a multi-task learning approach for DVOS through paired frame/mask prediction (as shown in Figure~\ref{fig:main_figure_01}D). Unlike conventional VOS methods, we exclude the query frame and reference masks from the model’s inputs to reduce reliance on explicit annotations (cf. Figure~\ref{fig:main_figure_01}C). Our design leverages the model’s ability to infer temporal dependencies from reference frames rather than directly exploiting the \textbf{Q}uery \textbf{F}rame and \textbf{R}eference \textbf{M}asks, as manipulated in~\cite{cheng2022xmem}, which we call \textit{QFRM-VOS} models. 
Using the query frame as an input reduces the model's reliance on capturing spatial and temporal information from the frames to predict the object's precise location in the subsequent frame, thereby improving mask prediction accuracy. In addition, under a linear assumption, removing the query frame from the list of model inputs reduces the computational cost by a factor of $\frac{1}{|\mathcal{R}| + 1}$, ignoring model-specific complexities. 
Furthermore, incorporating the first reference frame mask ($\mathcal{M}_{t - \tau}$) alongside $\mathcal{M}_t$ doubles the annotation requirements for training and validation. It also requires manual annotation of the first frame of each video during inference or prediction.  
Nevertheless, this effect is less pronounced in general-purpose tasks with large objects but becomes more significant in tasks involving videos of small objects. Our experiment shows that \textit{QFRM-VOS} models directly transfer the initial frame's mask without accounting for the frame's spatial pattern, indicating a tendency to rely on shortcuts rather than proper learning.
%
\subsection{Model Architecture}\label{subsec:DMArch}
\begin{figure}[!htb]
	\centering
	\includegraphics[width=\linewidth]{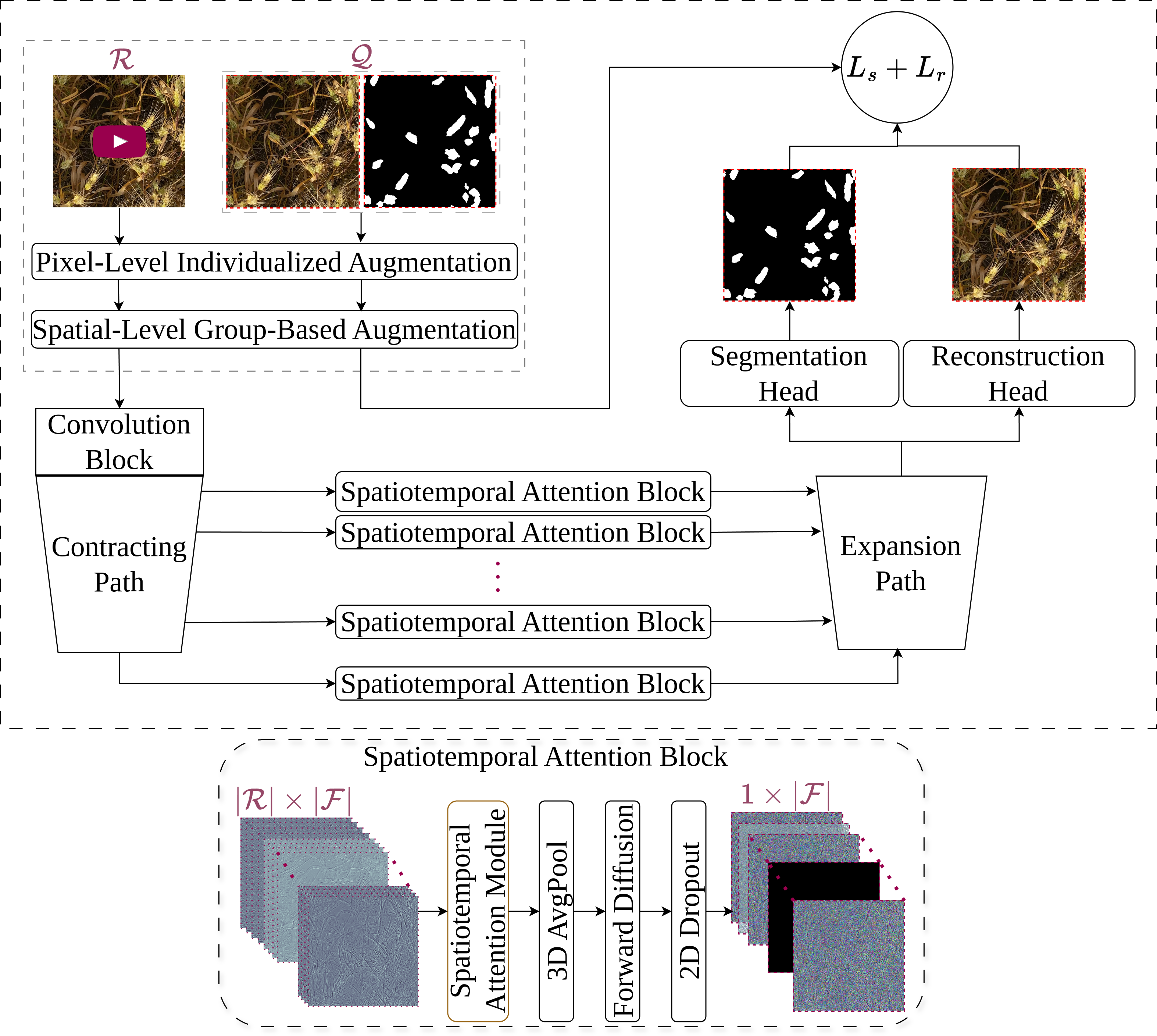}
	\caption{Overview of the proposed UNet-style~\cite{ronneberger2015u} architecture for DVOS. 
	}
	\label{fig:main_figure02}
\end{figure}

We designed a UNet-style~\cite{ronneberger2015u} 2D convolutional architecture for DVOS, trained with a multi-task learning paradigm. This model processes $\tau$ consecutive reference frames, $\mathcal{R}$, to predict the subsequent frame ($\mathcal{Q}$) and its mask ($\mathcal{M}$).
The key components of the proposed architecture include an initial convolution block, a contraction path, skip attentions, skip diffusion, an expansion path, and decoder heads, detailed below and visually illustrated in Figure~\ref{fig:main_figure02}.  \par 
The \textbf{initial Convolution Block} consists of a $3\times3$ convolution layer (stride 1, padding 1), followed by two ResNet blocks (Figure~\ref{fig:sub_figure03}). Each ResNet block includes two components, each with a Group Normalization~\cite{wu2018group} (size $8$), Swish activation~\cite{Ramachandran2017SwishAS}, and a $3\times3$ convolution with padding 1. This initial block preserves the input spatial dimension. \par 

The \textbf{Contracting Path} contains contractive modules, each with two residual blocks followed by a Group Normalization, activation, and $3\times3$ convolution (stride 2, padding 1) for spatial downsampling. The output is processed with \textbf{Spatiotemporal Attention Block} consisting of spatiotemporal attention, channel reduction, skip diffusion, and dropout. \par 

Inspired by~\cite{hu2018squeeze}, the spatiotemporal attention module aggregates informative signals from the concatenated $|\mathcal{R}| \times |\mathcal{F}|$ feature maps. It has two streams: a Spatial Attention stream using Depth-Wise Separable Convolutions~\cite{guo2019depthwise}, Group Normalization, Swish, and a Temporal Attention stream that combines $\tau$ feature maps with adaptive average pooling and two linear layers. This module preserves the input feature map dimensions ($B \times C \times H \times W$) but requires additional processing before entering the expansion path. Instead of reducing feature dimensions with DSC, we apply 3D Average Pooling, integrating spatiotemporal features. We also add noise through a diffusion-based scheduler and apply 2D Dropout to prevent over-reliance on skip connections. \par 

UNet models~\cite{ronneberger2015u} retain low- and high-level spatial information via skip connections, which enable precise localization but struggle with noisy or masked inputs. This limitation arises from their reliance on high-resolution feature maps, neglecting mid-level features~\cite{oktay2018attention}. To overcome this, we integrate a weighted forward diffusion process into the skip connections, balancing feature contributions and improving performance. Nevertheless, as opposed to generative diffusion models that focus on noise prediction, our approach implements reconstruction by utilizing low- and mid-level features. The diffusion process follows the Markov chain~\cite{ho2020denoising}, 
$x_t = \sqrt{\bar{\alpha_t}} x_0 + \sqrt{1 - \bar{\alpha}_t}\epsilon$, 
where $x_t$ is sampled from the diffusion kernel, specified as 
$q\left(x_t | x_0\right) = \mathcal{N}\left(x_t;\sqrt{\bar{\alpha}_t}x_0, \left(1 - \bar{\alpha}_t\right)\mathcal{I}\right)$. Here, 
$\epsilon \sim \mathcal{N}\left(0, \mathcal{I}\right)$, 
the scalar $\bar{\alpha}_t = \prod_{i=1}^{t}\left(1 - \beta_i\right)$, and 
$\beta_t$ serves as the variance scheduler, formulated to ensure that 
$\bar{\alpha}_T \rightarrow 0$. \par

In the latent space, no noise is applied initially, but noise increases along the expansion path, regulated by a scheduler based on beta distributions. The scheduler optimizes diffusion parameters across encoding/decoding levels (Appendix~\ref{app:beta_distro}, Figure~\ref{fig:sub_figure04}). \par 

The \textbf{Expansion Path} replicates the downsampling path, maintaining the number of residual blocks at each level. It uses in order Group Normalization, Swish, and Nearest-Neighbor upsampling (scaling factor 2) followed by a $3\times3$ convolution. Skip connections and lower-resolution feature maps are concatenated at each level. The shared decoder includes two heads: one for segmentation (1-channel output) and one for reconstruction (3-channel output). \par

Conventional UNet models often excel at reconstructing clear inputs but struggle with domain adaptation and generalization, particularly in segmentation and reconstruction tasks. To address this, we diffuse input images using a patching style (diffusion with $P_d = 0.5$, random time steps from 0 to 1000). Figure~\ref{fig:sub_figure05} shows an input image subjected to diffusion. We also apply pixel-level color augmentation, light color alteration, blur, and group-wise spatial transformations (small-angle rotation, random cropping, normalization) to both reference and query frames/masks during training. \par

%
\subsection{Data}\label{subsec:data}
We train the model in two phases, with the data section organized into two subsections that describe the datasets used for each phase. Furthermore, we introduce the test sets utilized for evaluating the developed models in the following subsection.

\subsubsection{Phase 1: Synthetic Data}
We generated a large-scale dataset of synthetic video sequences, denoted as $\mathbb{S}_{train}$, by using the cut-and-paste method introduced in~\cite{Najafian2021ASL, Najafian2022SemiSelfSupervisedLF}, to generate video sequences that emulate both camera motion (combined motion of background and foreground objects) and plant motion (motion of the foreground wheat heads relative to the background). Background frames are generated from video clips of bare fields (without wheat plants) by a cropping procedure that forms a sequence of consecutive frames as the background. Foreground wheat heads are extracted from seven annotated frames of wheat field videos. Fake wheat heads are included as negative-samples and are generated with the shape of real heads but with a non-head texture/appearance. \par

To simulate natural movement and deformation, the wheat heads underwent spatial and pixel-level transformations, mimicking both camera and environmental effects. This includes object-level movement (individual adjustments for each wheat head) and frame-level motion (adjustments for all wheat heads based on predefined behaviors). The frames and corresponding masks were synthesized simultaneously to maintain consistent annotations. Further details of the synthesis process can be found in Appendix~\ref{app:video_synthesis}, with visual illustrations in Figures~\ref{fig:sub_figure01} and~\ref{fig:sub_figure02}. \par

\begin{figure}[!ht]
    \centering
    \includegraphics[width=\linewidth]{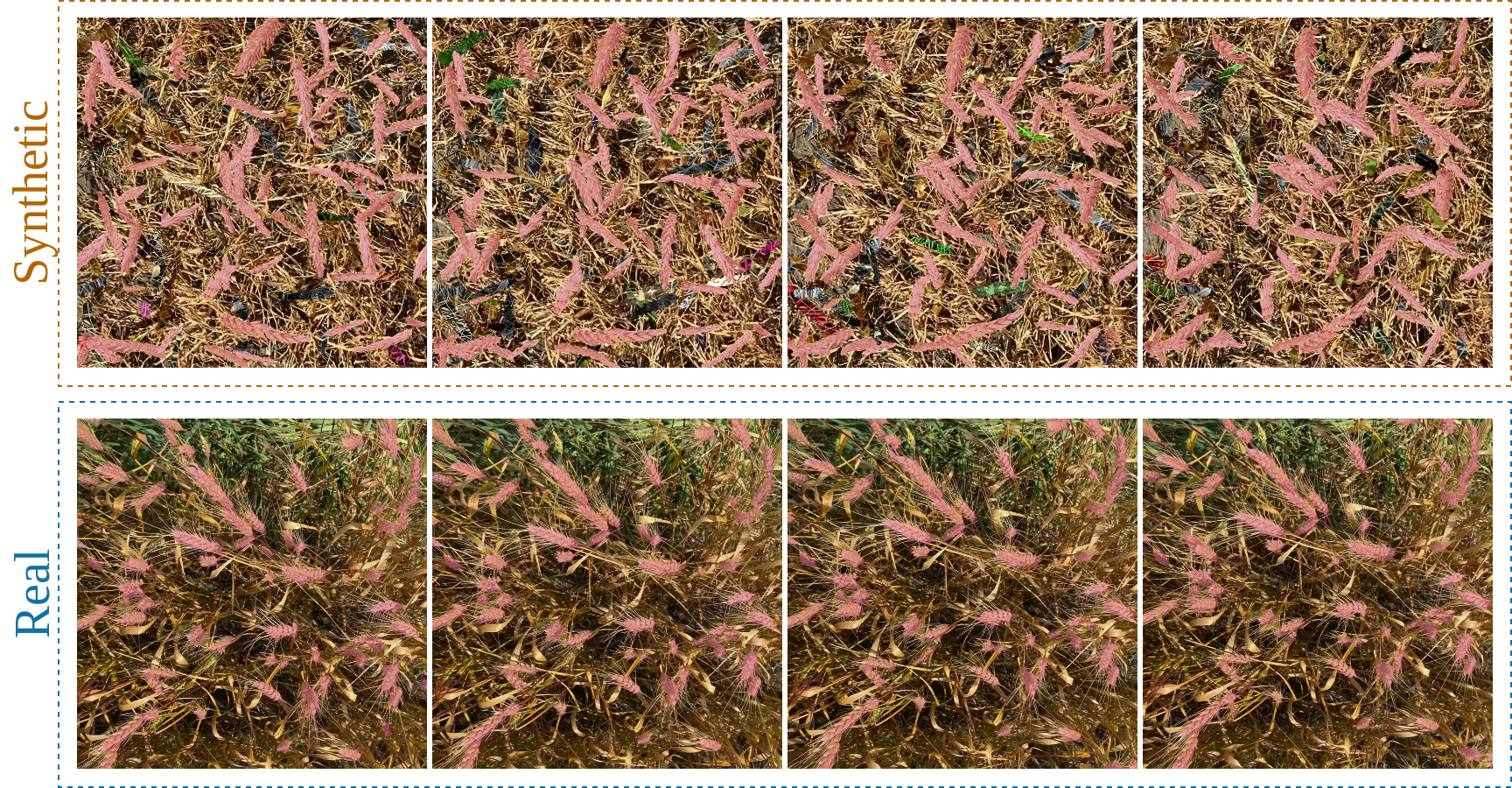}
    \caption{Synthetic videos show color-augmented fake wheat heads and masked real heads, isolated from the canopy and stem, overlaid on uniform background frames with random head-level movements. Real videos depict actual wheat fields, capturing normal motion in dense wheat spikes within the field.
    }
    \label{fig:main_figure03}
\end{figure}

The synthesized dataset, $\mathbb{S}_{train}$, consists of two subsets, Green Shaded and Yellow Shaded, each corresponding to different growth stages. The Green Shaded subset contains $13$ background videos with $101$ heads, resulting in $260$ synthesized longs videos and $15,600$ training samples, while the Yellow Shaded subset contains $15$ background videos with $251$ heads, resulting in $600$ synthesized videos and $36,000$ training examples. Figure~\ref{fig:main_figure03} shows a comparison between synthetic and real video frames, with masks overlaid in pink. \par
We also used a validation set of $300$ samples (denoted as $\ValidSingleDomainSamples$) including five consecutive video frames with the first four frames ($\RWindow{t}$) unannotated and the last one ($\QueryFrame{t}$) annotated. This validation set is from the same distribution of the Yellow Shaded subset of $\mathbb{S}_{train}$. The annotations were conducted in a semi-automated manner, where we manually refined the predictions made by our prior model from~\cite{Najafian2022SemiSelfSupervisedLF}. \par

\subsubsection{Phase 2: Pseudo-labeled Data}
We collected a set of $19$ top-view captured videos of wheat fields, denoted as $\WheatVideos=\{W_i\mid 1 \leq i \leq 19\}$, resulted in $86,572$ frames, representing various growth stages and environmental conditions. These videos were captured using a 12-megapixel camera from $1.0$, $1.5$, and $2.0$ meters altitudes. All videos
were resized from their original size of $2160 \times 3840$ pixels such that the height was scaled to $1024$ pixels while preserving the aspect ratio, then center-crops of size $1024 \times 1024$ were extracted from the resulting videos.\par
Since manual pixel-level annotation of such large-scale datasets is impractical, we utilized a model~\cite{Najafian2022SemiSelfSupervisedLF} built for wheat head image segmentation to automatically annotate the videos in $\mathbb{W}$ frame by frame. This resulted in computationally annotated video dataset $\PseudoLabeledVideos = \{(W_i, M_i) \mid 1 \leq i \leq 19\}$, which contains $86,572$ annotated images from these videos.
Figure~\ref{fig:main_figure04} illustrates sample image frames from these videos.\par 
A group-wise data split was conducted on $\PseudoLabeledVideos{}$ to partition it into training ($\PseudoLabeledVideos{train}$) with $7,525$ video clips and $37,613$ individual frames, validation ($\PseudoLabeledVideos{valid}$) with $2,839$ clips $14,182$ frames, and test set ($\PseudoLabeledVideos{test}$) of $6,958$ video clips and frame size of $34,777$. Then, longer videos were split into clips of five consecutive frames. This split approach ensured that data points from a single video clip were contained entirely within either training, validation, or test set while preventing overlap. When training with a frame interval of $\tau$, we merged the $\tau$ consecutive clips into one, ensuring that each merged clip originates from unique individual clips. Note that the test set was visually verified to ensure high data diversity, serving as a reliable external benchmark for evaluating the models' generalizability.

\subsubsection{Pixel-Accurate Manually Annotated Test Sets}
We used a manually annotated set (denoted as $\TestSingleDomain$) consisting of $100$ samples for testing. Each sample in $\TestSingleDomain$ included a combination of five consecutive image frames, where only the final frame ($\QueryFrame{t}$) was manually annotated. \par
\begin{figure}[!ht]
    \centering
    \includegraphics[width=\linewidth]{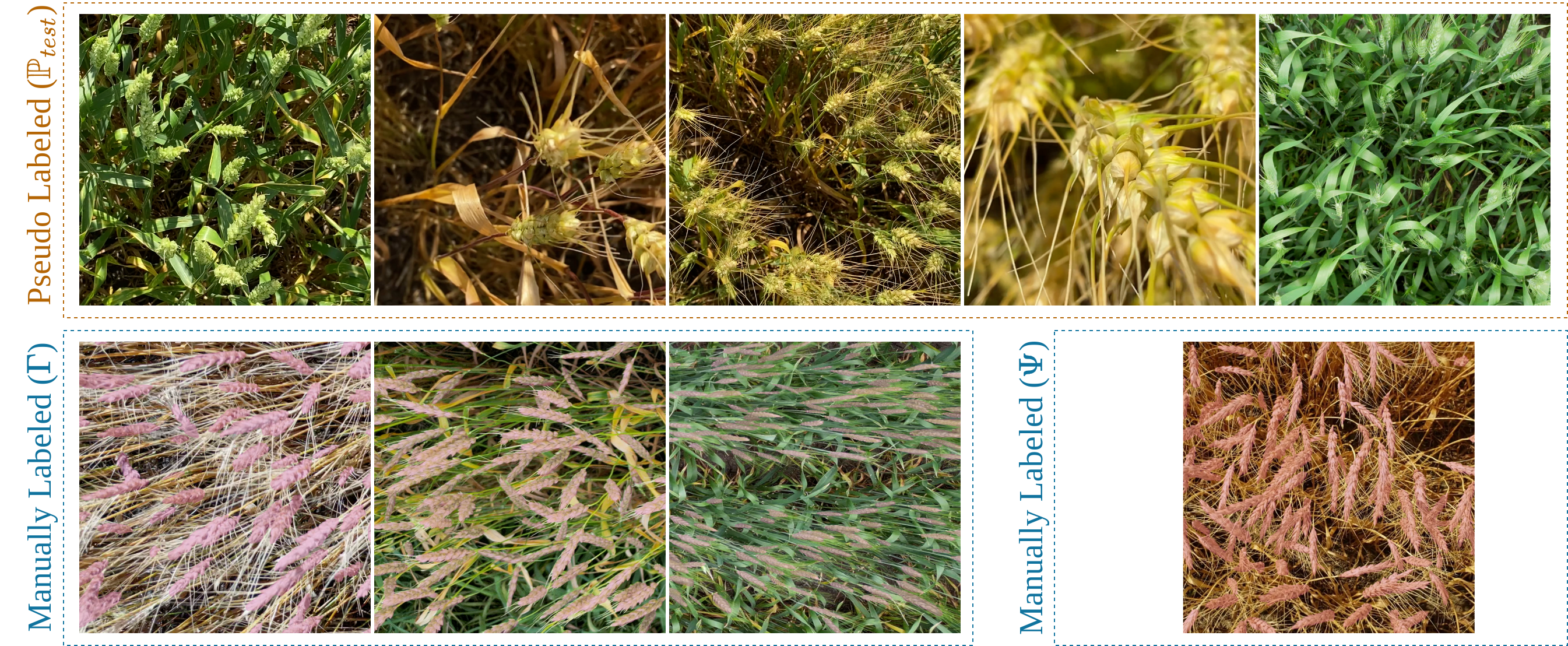}
    \caption{Representative examples of the test sets: the dashed orange box highlights the diversity of the pseudo-labeled dataset, and the blue boxes show manually annotated test set examples with overlaid annotations.}
    \label{fig:main_figure04}
\end{figure}

We also used an external test set, $\TestDroneSamples$, of $48$ samples, each containing five frames. The first four frames were unannotated, and the final frame was manually annotated. These samples were extracted from three distinctly diverse drone-captured videos of wheat fields to evaluate the model performance on the pixel-accurate annotated images. Each drone video contributed $16$ samples to $\TestDroneSamples$. The drone videos were captured using a \textit{DJI Mini 3 Pro Drone} from various altitudes. Importantly, $\TestDroneSamples$ is regarded as an external dataset since no samples from any of its domains are utilized to train or validate the models. Figure~\ref{fig:main_figure04} presents a visual representation of the test sets described in the preceding sections. \par

\subsection{Experiments}\label{subsec:development}
We trained all models identically in two phases for fair comparison. First, models were trained on $\mathbb{S}_{train}$ with $\Delta$ as the validation set. Then, they were fine-tuned on pseudo-labeled $\mathbb{P}_{train}$ and $\mathbb{P}_{valid}$, producing the final models. \par

Dice and IoU were used as segmentation metrics. For frame prediction ($L_r$ in Figure~\ref{fig:main_figure02}), we combined MSE and SSIM losses, while segmentation loss ($L_s$ in Figure~\ref{fig:main_figure02}) consisted of BCE and Dice. Given the dataset sizes, models were trained for $15$ epochs with batch sizes of $32$ and $16$ for image- and video-based models, respectively. The best model was selected based on the validation Dice score. Training used the AdamW optimizer~\cite{loshchilov2017decoupled} with a $1e-4$ learning rate and $1e-5$ weight decay, with CosineAnnealingLR gradually reducing the rate to $1e-5$ over $15$ epochs. \par

Training images were randomly cropped to a range from $256$ to $768$ out of the original $1024 \times 1024$ frames, resized to $384$. During the evaluation phases, a single $512$-pixel center crop ensured consistency. The architecture featured five downsampling modules with convolution kernels of $32$, $64$, $64$, $128$, $256$, and $512$ at the latent level, each with two residual blocks. Video models employed Group Normalization (group size = $4$) and a $0.3$ dropout in spatiotemporal modules. Middle convolution layers of the decoder heads used $32$ kernel sizes, while input/output layers matched decoder/task-required output dimensions. The model contained $\sim$92M parameters for video tasks and $\sim$16M for frame-based tasks, which utilized only encoder-decoder and head components. \par

To compare to image-based models, we removed the Spatiotemporal Attention Block (Figure~\ref{fig:main_figure02}), training with conventional skip connections instead, to develop \ImageSyntheticModel{} and \ImagePseudoModel{} as backbones for video models. Training followed two paradigms: (1) query frame as input (QAI) for a reconstruction/segmentation task, and (2) randomly selected reference frame as input (RRAI) for frame/mask prediction. In RRAI, UNet-style attention-free models predicted query frames and masks given a randomly chosen reference frame from $\mathcal{R}$ as input. \par

For DVOS-based models, we conducted multiple experiments either from scratch or leveraged frame-based models as partially pretrained models for DVOS. Our video-based models, \VideoSyntheticModel{} and \VideoPseudoModel{}, performances are thoroughly illustrated in section~\ref{sec:results} and supplementary section~\ref{sec:ablation_studies}. \par

Finally, we evaluated our models against XMem~\cite{cheng2022xmem}, which belongs to the state-of-the-art QFRM-VOS models. XMem processes reference frames, the query frame, and the initial reference mask as input to segment the query frame. To ensure fairness, XMem was trained with the same two-phase procedure as our models and compared to our best-performing DVOS model. The trained models are referred to as~\XMemSyntheticModel and~\XMemPseudoModel{}. \par
\section{Results}\label{sec:results}
In this section, we present the performance of our best-performing model (\VideoSyntheticModel{*}, \VideoPseudoModel{*} with asterisk superscript in Table~\ref{table:ablation_results}) alongside the reference models, \XMemSyntheticModel and \XMemPseudoModel. For simplicity, we remove the asterisk superscript here. Additional experiments and ablations are provided in Section~\ref{sec:ablation_studies} in the supplementary materials. \par 
\begin{table}[!th]
\centering
\caption{Quantitative evaluation on manually annotated pixel-accurate test sets: handheld videos ($\Psi$) and drone-captured ($\Gamma$). \VideoSyntheticModel{}, partially pretrained on a QAI frame-based task, and \XMemSyntheticModel{} trained on $\mathbb{S}_{train}$ and $\Delta$. Additionally, \VideoPseudoModel{} and \XMemPseudoModel{} were fine-tuned on $\PseudoLabeledVideos{train}$ and $\PseudoLabeledVideos{valid}$.
}
\label{tab:table_03}
\begin{tabular}{ccccccc}
\hline
\textbf{Model} & \textbf{Pretrained On} & \textbf{Metric} & \textbf{$\Psi$} &  \textbf{$\Gamma$} \\ \hline
\multirow{2}{*}{\ImagePseudoModel{}} & \multirow{2}{*}{\ImageSyntheticModel{}} 
                & Dice & 0.759 & 0.761  \\
               && IoU  & 0.621 & 0.619  \\ \hline
\multirow{2}{*}{\VideoSyntheticModel{}} & \multirow{2}{*}{None} 
                & Dice  & 0.482 & 0.453  \\
               && IoU & 0.335 & 0.307  \\ \cline{2-5}
\multirow{2}{*}{\VideoPseudoModel{}} & \multirow{2}{*}{\VideoSyntheticModel{}} 
        &  Dice  & \textbf{0.650} & \textbf{0.791}  \\
        && IoU   & \textbf{0.493} & \textbf{0.657}  \\ \hline
\multirow{2}{*}{\XMemSyntheticModel} & \multirow{2}{*}{XMem-s012~\cite{cheng2022xmem}}
        &  Dice  & 0.794 & 0.454  \\
        && IoU   & 0.668 & 0.314  \\ \cline{2-5}
\multirow{2}{*}{\XMemPseudoModel} & \multirow{2}{*}{\XMemSyntheticModel} 
        &  Dice  & 0.831 & 0.811  \\
        && IoU   & 0.716 & 0.690  \\ \hline
\end{tabular}
\end{table}
\begin{figure}[!ht]
	\centering
	\includegraphics[width=\linewidth]{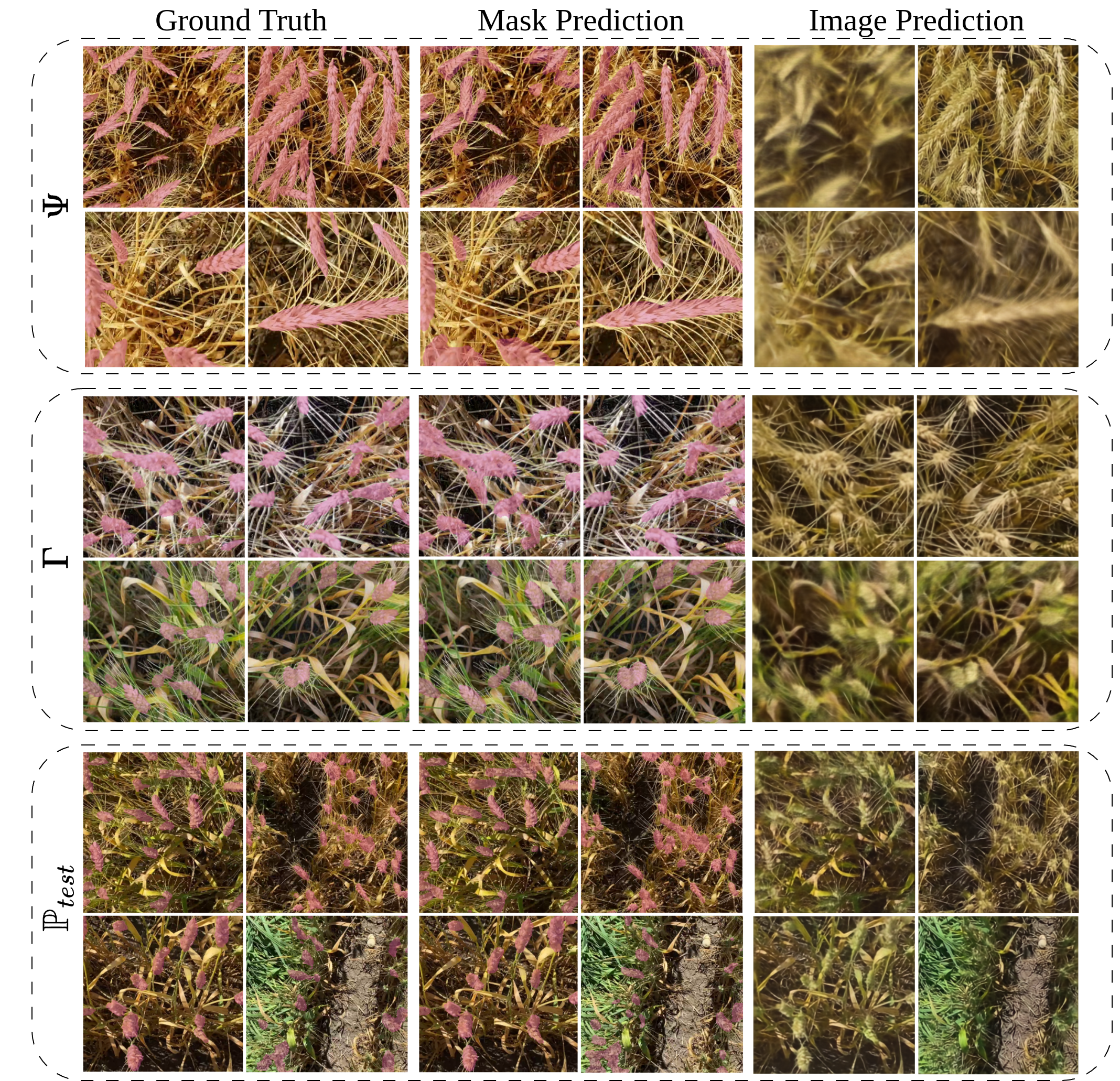}
	\caption{
		Performance visualization of the \VideoPseudoModel{} model across different test sets. The first two columns depict masks overlaid on the corresponding images. Each block forms four different samples, which are consistently arranged within the same grid cell across Ground Truth, Mask Prediction, and Image Prediction columns.
	}
	\label{fig:main_figure05}
\end{figure}
Table~\ref{tab:table_03} demonstrates the performance of our developed models across our manually annotated test sets.
Both image level \ImageSyntheticModel{} and \ImagePseudoModel{} were trained on individual frames of synthetic and pseudo-labeled datasets in the QAI paradigm. The \ImagePseudoModel{} model was evaluated at the frame level, demonstrating highly consistent performances across our manually annotated test sets (please see Section~\ref{sec:ablation_studies} for more details). \par 

\VideoSyntheticModel{}, which partially pretrained on the \ImagePseudoModel{}, trained on $\mathbb{S}_{train}$ data. Although it was trained exclusively on a synthetic dataset with precise annotations, its performance across both datasets is reasonable, considering the characteristics of the training data. Although the \VideoSyntheticModel{} model trained on the synthetic dataset, it demonstrated comparable dice scores of $0.482$ on $\Psi$ and $0.453$ on $\Gamma$. Note that the synthetic dataset was generated using only a limited selection of wheat heads from a few image frames, spanning two domains. In contrast, $\Psi$, which was captured at a $30$-fps configuration, serves as a single-domain real-field evaluation set. Meanwhile, the dataset $\Gamma$ serves as an external test set for this model and includes videos captured from varying altitudes, under different weather conditions, and with various light intensities. \par

Our fine-tuned model, \VideoPseudoModel{}, exhibits notably improved performance when trained on the large-scale and diverse pseudo-labeled dataset, $\mathbb{P}_{train}$. It achieved a substantial performance gain of over $16.8\%$ on the $\Psi$ dataset, despite lacking training samples from this domain, and nearly $43.5\%$ on $\mathbb{P}_{test}$ (Table~\ref{tab:table_04}). Furthermore, the model's accuracy on the manually labeled dataset underscores its precision in pixel-accurate segmentation of wheat head objects. This model also demonstrates significantly enhanced performance when evaluated on $\Gamma$, achieving a Dice score of $0.791$. This represents an improvement of over $33\%$ compared to model \VideoSyntheticModel{}. \par

Figure~\ref{fig:main_figure05} visually illustrates the prediction performance of model \VideoPseudoModel{} on randomly selected samples from the test sets. A grid format is used to arrange four randomly selected samples in each test set. 
The left grids display the ground truth masks overlaid on their corresponding images. The middle grids present the model's mask prediction performance, visualized as the masks overlaid on the images. 
Despite the model's ability to reconstruct the query frame (third column grids) from noisy reference frames, diffused by input and hierarchically added skip-level noise, we solely utilized the model's frame prediction capability to achieve more stable and effective training. It is important to note that the ground truth for the second row of grids represents the model-generated (pseudo) masks. \par

\begin{figure}[!ht]
\centering
\includegraphics[width=\linewidth]{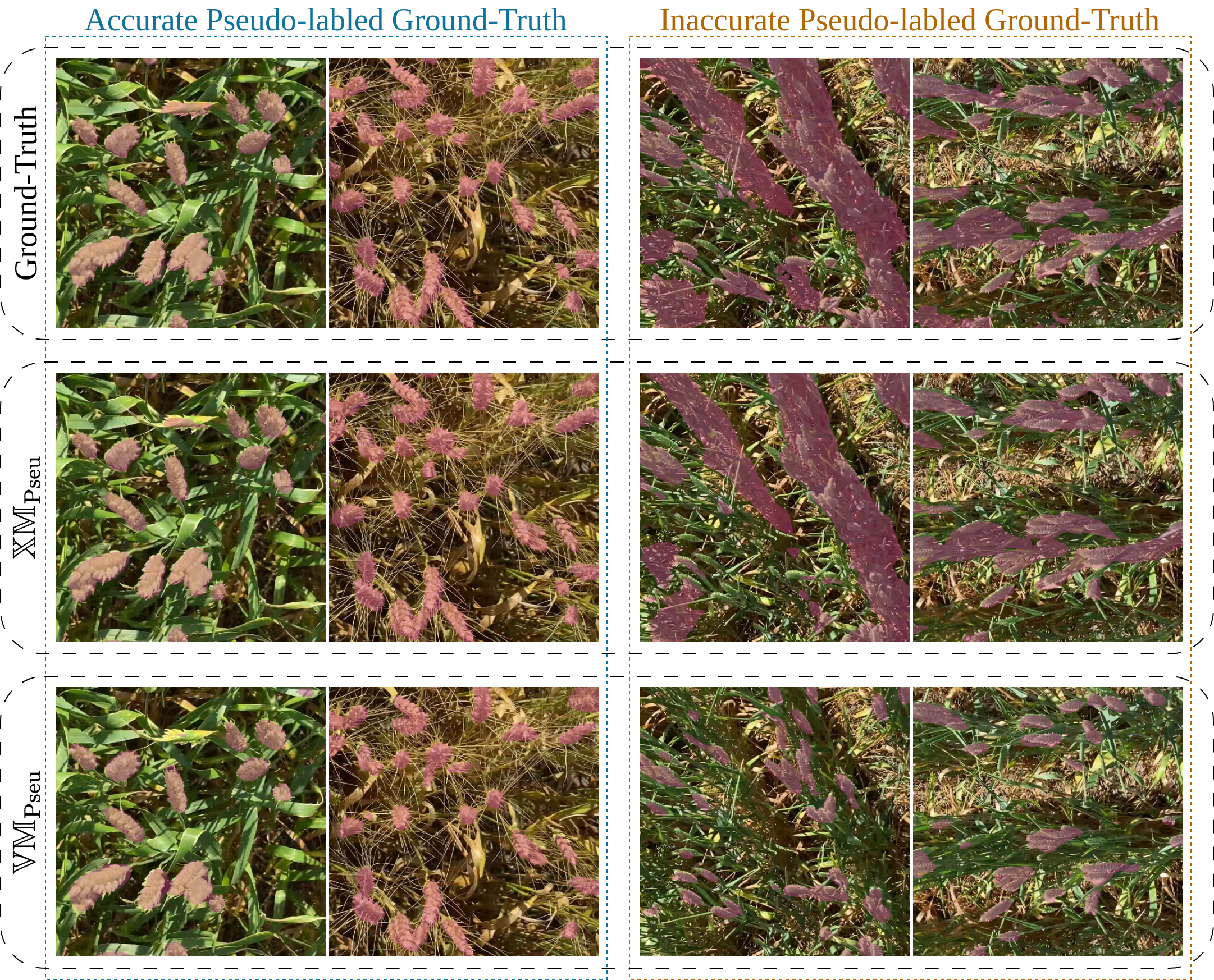}
\caption{Visual inspection of model prediction performances: The left columns, in dashed blue boxes, demonstrate high-quality predictions by both models when the pseudo-labeled ground truth is pixel-accurate. The right columns, in orange boxes, illustrate the degradation in \XMemPseudoModel performance due to its reliance on the initial reference frame, in contrast to our \VideoPseudoModel{} model, which remains robust even in the presence of inaccurate ground truth.
}
\label{fig:main_figure06}
\end{figure}

We compare our approach to the XMem~\cite{cheng2022xmem} model, which represents an upper-bound on VOS performance because it uses the first-frame mask as input together with the reference frames and the query frame itself (similar to Figure~\ref{fig:main_figure_01}C) whereas our method excludes query frame and input masks entirely and relies solely on the reference frames (Figure~\ref{fig:main_figure_01}D). The comparison across all three test sets is presented in Tables~\ref{tab:table_03} and~\ref{tab:table_04}. The quantitative results indicate that XMem consistently achieves overall Dice scores above $0.8$ on all three test sets, and our best-performing model reaches just below this level of performance. In addition, the qualitative segmentation results of \XMemPseudoModel, shown in Figure~\ref{fig:main_figure06}, highlight some problems that are overcome by our approach, such as:
\begin{itemize}[leftmargin=1em]
    \item The XMem model generates segmentation masks for subsequent frames based on the provided initial frame's mask. Consequently, when the initial frame mask is highly accurate, the model produces near-perfect predictions (as illustrated by the dashed blue box in Figure~\ref{fig:main_figure06}). Conversely, when the initial mask is inaccurate, the segmentation quality remains poor or further degrades over subsequent frames (dashed orange box in Figure~\ref{fig:main_figure06}). This dependency results in high quantitative scores when the subsequent ground-truth masks closely resemble the initial reference frame mask.
    \item The model exhibits a strong dependency on the initial reference frame's ground-truth mask, limiting its ability to effectively learn spatial and temporal information from individual frames. Rather than adapting to the content of each frame, it primarily propagates the provided mask throughout the sequence in a cascading manner, leading to ineffective learning. 
 \end{itemize}

This is in contrast to our \VideoPseudoModel{} model, which despite achieving lower quantitative scores, effectively identifies and segments the actual wheat objects in both scenarios, demonstrating robustness in segmentation performance regardless of the ground-truth accuracy. \par

\begin{table*}[!th]
\centering
\caption{Quantitative results of models \VideoSyntheticModel{}, \VideoPseudoModel{} and \XMemPseudoModel on individual videos from $\PseudoLabeledVideos{test}$ along with the overall average score, weighted by the number of frames rather than a simple balanced average across the five videos.
}
\label{tab:table_04}
\begin{tabular}{ccccccccc}
\hline
\textbf{Model} & \textbf{Trained On} & \textbf{Metric} & \textbf{Video 1} & \textbf{Video 2} & \textbf{Video 3} & \textbf{Video 4} & \textbf{Video 5} & \textbf{All}\\ \hline
\multirow{2}{*}{\VideoSyntheticModel{}} & \multirow{2}{*}{\ImagePseudoModel{}}                           
         & Dice    & 0.160 & 0.281 & 0.219 & 0.204 & 0.314 & 0.244  \\
        && IoU     & 0.099 & 0.174 & 0.130 & 0.117 & 0.207 & 0.150  \\ \cline{2-9}

\multirow{2}{*}{\VideoPseudoModel{}}     & \multirow{2}{*}{\VideoSyntheticModel{}} 
         & Dice    & 0.480 & 0.710 & 0.825 & 0.821 & 0.820 & 0.679  \\
        && IoU     & 0.330 & 0.571 & 0.705 & 0.699 & 0.712 & 0.542  \\ \hline

\multirow{2}{*}{\XMemPseudoModel}     & \multirow{2}{*}{\XMemSyntheticModel} 
         & Dice    & 0.731 & 0.845 & 0.932 & 0.918 & 0.934 & 0.835  \\
        && IoU     & 0.579 & 0.736 & 0.873 & 0.848 & 0.878 & 0.726  \\ \hline
\end{tabular}
\end{table*}
We also evaluated the \VideoPseudoModel{} model performance on each video $\mathbb{V} \in \mathbb{P}_{test}$, where each video captures a diverse exhibition of wheat plants, collectively representing diverse phenotypes, often from slightly variant altitudes. While all models demonstrate consistent performance across most videos, \textit{Video 1} proves more challenging to segment compared to others (see dashed orange columns in Figure~\ref{fig:main_figure06} and Table~\ref{tab:table_04}). This difficulty is evident in the performance of \VideoSyntheticModel{}, \VideoPseudoModel{}, and \XMemPseudoModel models. As discussed above, the presence of low-quality video results in impaired pseudo-labeled ground truth, thereby diminishing the quantitative scores of our models. Despite this, visual inspection reveals proper segmentation of \textit{Video 1} by our model, capturing even the low-quality wheat heads. \par

\section{Discussion}\label{sec:discussion}
In this paper, we proposed a semi-supervised method for segmenting videos with dense patterns using only seven manually annotated frames for large-scale data synthesis, weakly-supervised learning with pseudo-labels, and self-supervised learning via reconstruction. By leveraging synthesized and pseudo-labeled clips, we bypassed the costly manual labeling step typically required for videos of small and dense objects, such as videos of crop fields. \par

We engineered an automatic data synthesis method, replacing manual annotation by simulating motion and optical flow. Extending~\cite{Najafian2022SemiSelfSupervisedLF} with object- and frame-level motions, we generated diverse datasets with various backgrounds. The synthetic data alone achieved a Dice score of $0.42$ to $0.48$ with \VideoSyntheticModel{} when evaluated on our manually-annotated test set (Table~\ref{tab:table_03}). Fine-tuning with a few minutes of pseudo-labeled wheat-field videos further improved the model's performance by $17\%$ to $43\%$ across different test sets. \par

Furthermore, we introduced a convolution-based architecture for DVOS tasks, integrating a hierarchical diffusion process and a spatiotemporal mechanism via multi-task frame/mask prediction. Retaining the well-established UNet-style~\cite{ronneberger2015u,oktay2018attention} architecture with shared encoders and decoders, we added two task-distinct heads for frame and mask prediction and incorporated spatiotemporal attention modules into the skip connections to effectively capture and temporal dependencies across frames. \par

In contrast to existing QFRM-VOS approaches~\cite{dai2016instance, maninis2018video, meinhardt2020make, cheng2022xmem, cheng2021rethinking}---requiring initial frame ground truth, reference, and query frames as input for training and even inference---our model used only reference frames, reducing manual annotation by half and mitigating initial mask dependency.
Further, we showed that our training approach prevented models from exploiting initial masks via the waterfall effect as happened to XMem models (Figure~\ref{fig:main_figure06}). \par

Our flexible architecture enabled training on small-scale data. Without manual annotations, \VideoPseudoModel{} trained on synthetic and pseudo-labeled clips achieved comparable performance to XMem~\cite{cheng2022xmem}, obtaining a $0.79$ dice score on the drone-captured test set, $\Gamma$.
Furthermore, our approach proved significantly more accurate when evaluated on the pseudo-labeled data with the inaccurate initial frame’s mask (dashed orange box in Figure~\ref{fig:main_figure06}), reinforcing the reliance of existing methods on accurate initial segmentation and highlighting a key limitation of QFRM-approaches (Figure~\ref{fig:main_figure06}). \par

We demonstrated that our integrated diffusion-based and augmentation preprocessing and skip diffusion model components work effectively. UNet-style models transfer unprocessed high-level features via skip connections~\cite{oktay2018attention}, which can affect segmentation in dense object scenarios. To fix this, we integrated a diffusion-based hierarchical module into our spatiotemporal module on each skip connection to regulate the internal information flow, applied heavy preprocessing pixel-level augmentation and tile-based diffusion noise, aiming at enhancing model robustness to videos' inherent noise and environment-enforced noise. Further, introducing additional complexity during training, increases task difficulty through controlled perturbations, aiming to act as an implicit regularization mechanism and reduce overfitting risks. \par

\VideoPseudoModel{}, trained only on synthetic data and fine-tuned on pseudo-labeled data, accurately segmented wheat heads (Figure~\ref{fig:main_figure06} and supplementary Figure~\ref{fig:sub_figure06}), demonstrating robustness in disregarding misleading segmentation signals. Video 1's lower performance highlights video quality's impact on model evaluation and deployment. However, visual inspection confirms that our developed model performs well across all five videos, outperforming the ground truth and \XMemPseudoModel in segmenting Videos $1$ and $2$, where the ground truth lacks pixel-accurate annotation. \par

To enhance performance, we propose expanding human-annotated images for foreground object extraction ($\mathbb{H}$, $\overline{\mathbb{H}}$ in supplementary section~\ref{app:video_synthesis}), extracting wheat heads from individual images instead of noisy video frames, and improving synthetic data quality. We also observed that temporal consistency among training video clips is crucial for deep learning models (supplementary section~\ref{sec:ablation_studies}). Future work could synchronize wheat head movements across frames for more consistent and realistic motion, aligning with natural dynamics in synthesizing agricultural and general-purpose videos. The second phase of model training used authentic, pseudo-labeled videos without pixel-level annotations, mainly consisting of long video clips. We recommend expanding the dataset with shorter, more phenotypically diverse wheat video clips. \par

\section{Conclusion}\label{sec:conclusion}
To conclude, we have proposed a semi-self-supervised strategy for dense video object segmentation, addressing annotation challenges for small and occluded objects. We trained a novel multi-task learning architecture using synthetic videos and weakly-labeled real videos of wheat fields. Our model demonstrated high performance across diverse test sets despite limited human-annotated frames. The methodology can be extended beyond precision agriculture as it is applicable to video-centric tasks in autonomous driving and medical imaging requiring multi-object and dense pattern segmentation.
{
    \small
    \bibliographystyle{ieeenat_fullname}
    \bibliography{main}
}

\setcounter{page}{1}
\clearpage  
\onecolumn  

\renewcommand{\thesection}{S\arabic{section}}

\renewcommand{\thefigure}{S\arabic{figure}}

\renewcommand{\thetable}{S\arabic{table}}

\begin{center}
   \Large
   \textbf{\thetitle} \\  
   \vspace{1.0em}
   Supplementary Material \\  
   \vspace{1.0em}
   \author{Keyhan Najafian$^1$, Farhad Maleki$^2$, Lingling Jin$^1$, Ian Stavness$^1$\\
    {\large $^1$Department of Computer Science, University of Saskatchewan, Saskatoon, Saskatchewan, Canada} \\
    {\large $^2$Department of Computer Science, University of Calgary, Calgary, Alberta, Canada} \\
    {\tt\small \{keyhan.najafian, lingling.jin, ian.stavness\}@usask.ca, farhad.maleki1@ucalgary.ca}
    }
   \vspace{1.0em}
\end{center}

\section{Data Synthesis Procedure}\label{app:video_synthesis} 
We utilized video clips of wheat fields and background fields, which were fields without wheat.
An annotated video clip $\Video{i}=\{(x_{i_t}, y_{i_t})\}_{i_t=1}^{T_i}$, is defined as a series of $T_i$ consecutive image frames, where $x_{i_t}$ represents the ${i_t}^{th}$ frame of $V_i$, and $y_{i_t}$ represents the pixel-level annotation (mask) for $x_{i_t}$, representing wheat heads in $x_{i_t}$. \par

We used three top-view wheat field videos, $\mathbb{V}=\{V_1, V_2, V_3\}$, and $28$ background videos, $\mathbb{B}=\{B_i \mid 1\leq i \leq 28\}$, featuring fields without wheat. All videos were captured with a 12-megapixel handheld camera. The background videos contained $118,259$ frames. We manually annotated seven randomly selected frames from the wheat field videos, $\SynFrames=\{F_i \mid 1 \leq i \leq 7\}$, for data synthesis. Using these frames and extracted wheat heads, we applied a cut-and-paste strategy to overlay wheat heads onto background frames, generating computationally annotated video clips, thereby reducing manual annotation effort. \par

\begin{figure}[!ht]
    \centering
    \includegraphics[width=0.8\textwidth]{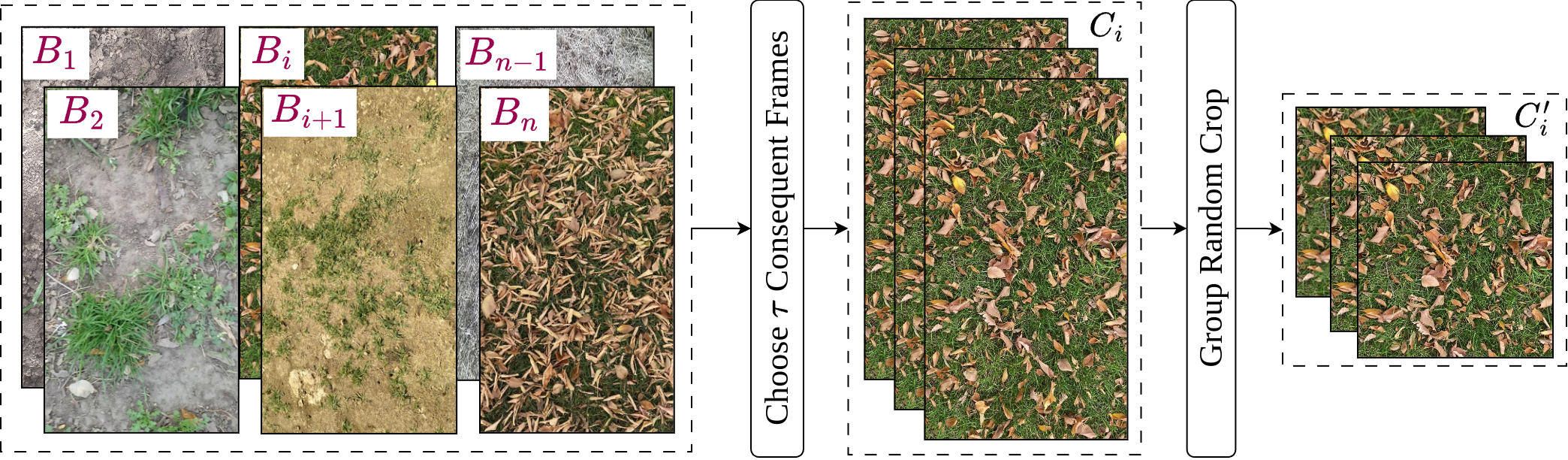}
    \caption{The procedure for extracting video frames from background video $B_i \in \mathbb{B}$.}
    \label{fig:sub_figure01}
\end{figure}

We employed a group of background videos ($\mathbb{B}$), which were fields without wheat crops exhibiting various vegetation types and environmental conditions, used as the backgrounds of synthetic videos. Figure~\ref{fig:sub_figure01} illustrates the background video frames extraction process. This was achieved by randomly choosing video $B_i$ from the background videos $\mathbb{B}=\{B_i \mid 1\leq i \leq 28\}$. Then, $\tau$ consequent frames from $B_i$ were selected to form a set $C_i =\{c_{i_t} \mid 1 \leq t \leq T_i - \tau\}$, where $T_i$ is the $B_i$'s length. Next, a random region of size $1024 \times 1024$ was chosen, and the crop defined by this region was applied to all the frames in $B_i$, resulting in a set of $\tau$ consecutive frames of size $1024 \times 1024$, denoted as $C_i^{\prime}$. This set was subsequently utilized as the background frames for overlaying objects of interest in synthesizing a video clip forming $\tau$ frames.\par
We also extracted wheat heads from a small number of manually annotated frames depicting both earlier growth stages in green shades and harvestable-ready stages in yellow coloration, resulting in a set of real wheat heads ($\mathbb{H}$) consisting of yellow (mature) and green (mid-season)  wheat heads. Note that, for wheat head extraction, we chose the frames from three distinct videos in $\mathbb{V}$ that had no intersection with the $\WheatVideos$ dataset used in the pseudo-labeling phase of model training. We also used extracted wheat heads in $\mathbb{H}$ as cookie-cutters to extract regions with no wheat heads (fake wheat heads) from the original frames, denoted as $\mathbb{\overline{H}}$.
\par
\begin{figure}[!ht]
    \centering
    \includegraphics[width=0.7\textwidth]{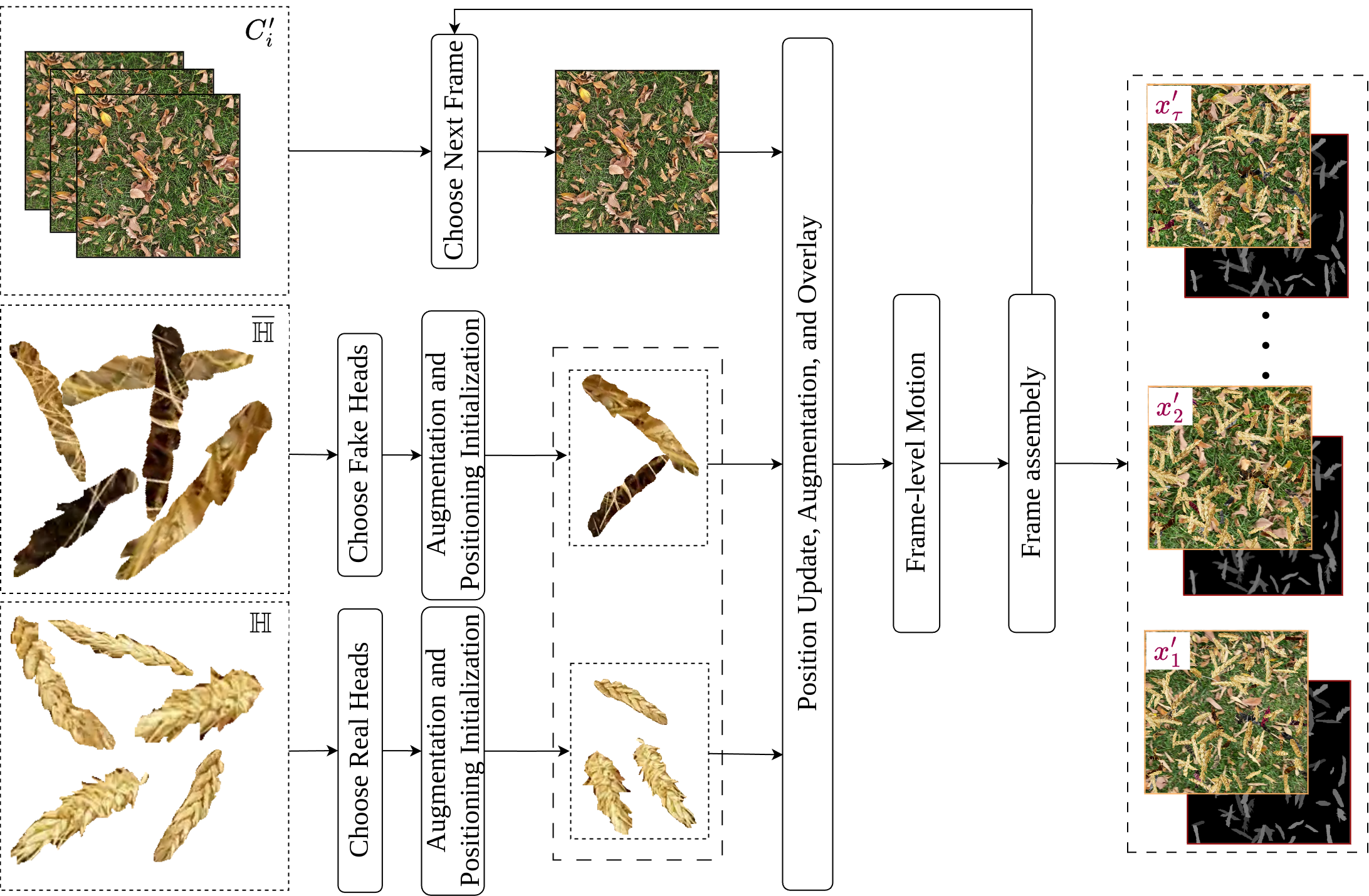}
    \caption{This diagram illustrates the process of synthesizing videos with dimensions of $1024 \times 1024$ and a length of $\tau$. We generated a synthesized video $V'$ by initializing a random selection of real and fake heads from $\mathbb{H}$ and $\mathbb{\overline{H}}$, thereby uniquely augmented and positioned by predefined parameters and overlaid on the first frame of $\tau$ frames in $C_i^\prime$. Subsequent frames of $V'$ were simulated based on the positions of wheat heads in the preceding frame, utilizing both object-level and frame-level motions. Objects that were not within the frame anymore (due to the object motions) were subsequently restored by incorporating additional real heads into the chosen object set and overlaid on the current frame before proceeding to the next frame. The frame masks were also generated simultaneously by applying the same object- and frame-level motions on the segmentation counters.
    }
    \label{fig:sub_figure02}
\end{figure}

 
Figure~\ref{fig:sub_figure02} illustrates the process for synthesizing a manually annotated video clip. For $\tau$ consecutive background frames in $C_i^{\prime}$, a random number of fake and real wheat heads were chosen randomly from $\mathbb{\overline{H}}$ and $\mathbb{H}$, respectively. The fake wheat heads were overlaid on the first frame of $C_i^{\prime}$, followed by real wheat heads. To simulate the movement and deformation of the crop naturally caused by wind, we applied a sequence of spatial- and pixel-level transformations to each wheat head before overlying them on consecutive frames in $C_i^{\prime}$. These transformations were automatically generated for each wheat head to ensure consistent movement while showing varying degrees of deformations.
Specifically, we defined two types of movement for each wheat head object: object-level and frame-level movement. In object-level movement, we updated the position and direction of each object individually. However, at the frame-level motion, all the objects' positions were adjusted according to a predefined motion behavior at the frame level. \par

This process synthesized frames and their corresponding masks by applying special transformations and deformations to both the frames and masks, ensuring consistent annotation for the resulting video clip. 
To create the mask corresponding to each synthesized video frame, a $1024 \times 1024$ blank frame was allocated for each of the $\tau$ frames in $C_i^{\prime}$. We kept track of the position of each real wheat head when overlaid on the background image and its movement and deformation to convert the corresponding region on the mask to $1$. Note that the fake wheat heads were ignored. Table~\ref{tab:table_02} provides a summary of the synthesized dataset, including statistical information on the synthesized videos and the raw data used in the synthesis process. This includes the number of background videos and the number of real wheat heads ($\mathbb{H}$). \par
\begin{table}[!ht]
\centering
\renewcommand{\arraystretch}{1.3}
    \begin{tabular}{cccccc}
        \textbf{Dataset}                    & \textbf{Subset} & \textbf{Background Videos} & \textbf{Heads} & \textbf{Synthesized Videos} & \textbf{Video Clips} \\ \hline \hline
        \multirow{2}{*}{$\mathbb{S}_{train}$} & Green Shaded    & 13                            & 101                     & 260                      & 15600                    \\
                                            & Yellow Shaded   & 15                            & 251                     & 600                      & 36000                    \\ \hline
    \end{tabular}
\caption{Quantitative summary of the synthetic videos and their frames distributions in the $\mathbb{S}_{train}$ dataset.}
\label{tab:table_02}
\end{table}

%
\section{Proposed Model Architecture Components}\label{app:beta_distro} 
This section presents the two key components of the proposed UNet-style architecture: the Residual Building Block (top dashed box) and the Spatiotemporal Attention Module (bottom dashed box). \par 

The Residual Building Block serves as the fundamental unit of our architecture. It comprises two sequential series of Group Normalization, Swish activation, and Convolutional layers, designed to enhance feature representation while preserving gradient flow through residual connections. Moreover, the layer arrangement within the Residual Building Block defines a standardized ordering framework for all preceding and succeeding modules and components in the architecture. \par

We also enhance the skip connections with a Spatiotemporal Attention Module, which operates alongside the diffusion module and feature reduction modules. This module enables the aggregation of feature maps from multiple input reference frames into a unified feature representation before being passed to the decoder. The Spatiotemporal Attention Module consists of two parallel processing streams: Spatial Attention (top) and Temporal Attention (bottom). \par 

The Spatial Attention Stream employs depth-wise separable convolutions, group normalization, and Swish activation to highlight informative spatial regions within individual frames. The Temporal Attention Stream captures dependencies across frames using adaptive average pooling, linear transformations, and layer normalization, followed by Swish or sigmoid activation to generate temporal attention weights. The final attention-enhanced feature maps are computed by element-wise multiplication of the original feature maps with the spatial and temporal attention outputs. This mechanism enables the model to effectively capture both spatial and temporal information, improving segmentation performance. \par

\begin{figure}
	\centering
	\includegraphics[width=0.7\textwidth]{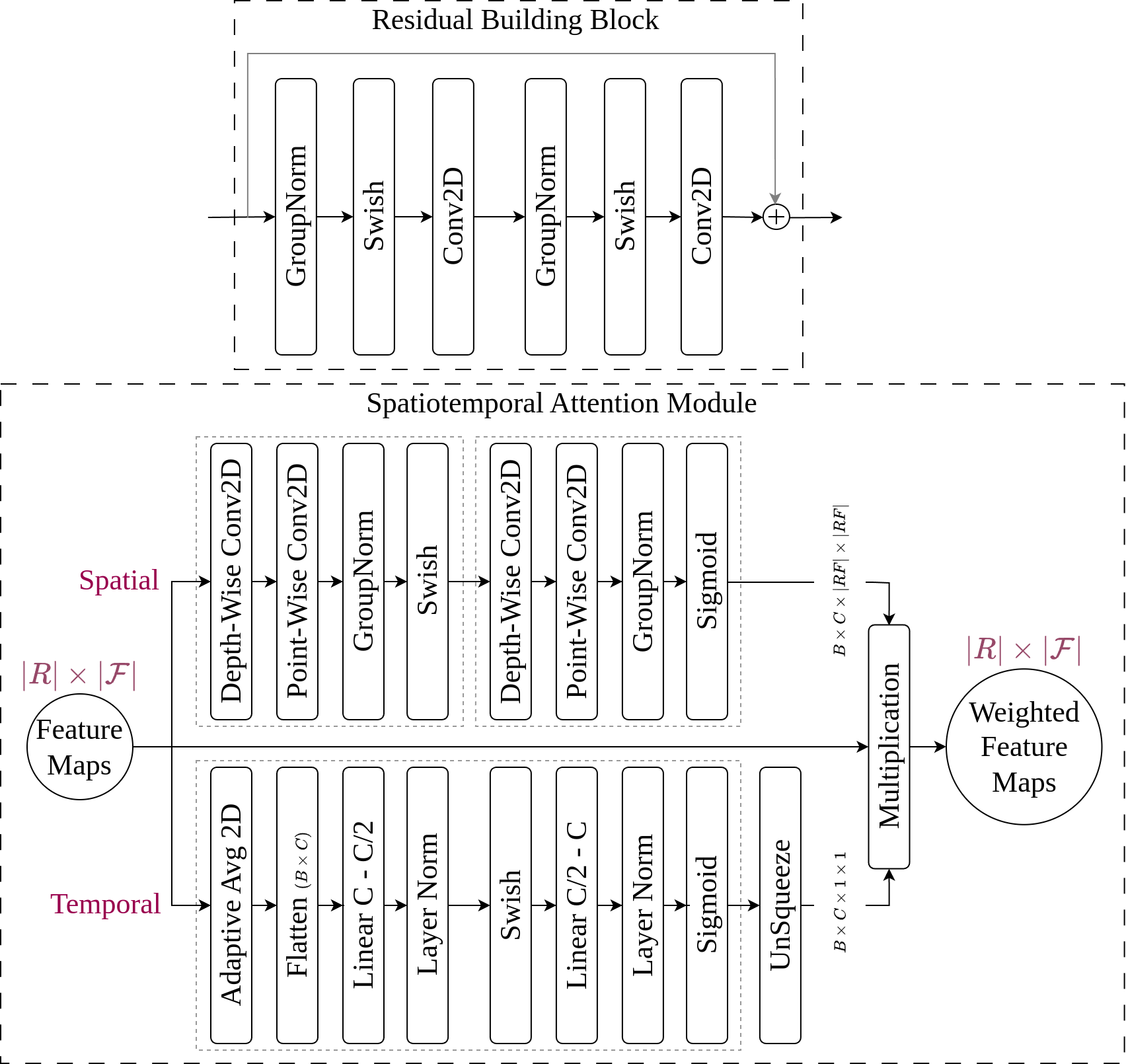}
	\caption{Top dashed-box: The residual building block represents the base component of the model architecture. Bottom dashed-box: The spatiotemporal attention module incorporates two processing streams for input feature maps: (1) Spatial Attention Stream (Top) captures and weights informative spatial regions within individual reference frames. Group normalization is employed to highlight the spatial importance of feature maps corresponding to each input frame. Additionally, depth-wise separable convolutions are used for the efficient processing of large-scale feature maps; (2) the Temporal Attention Stream (Bottom) focuses on capturing temporal dynamics and dependencies. It weights each time step across feature maps extracted from all consecutive frames.
	}
	\label{fig:sub_figure03}
\end{figure}
%

\section{The Beta Distribution Scheduler}\label{app:beta_distro} 
The beta distribution is parameterized by two positive shape parameters, denoted by $\alpha$ and $\beta$.
\begin{equation}
	f(x, \alpha, \beta) = \frac{1}{B\left(\alpha, \beta\right)} x^{\alpha - 1}\left(1 - x\right)^{\beta - 1}
\end{equation}
where $x \in [0, 1]$, and $B\left(\alpha, \beta\right)$ is the beta function, which is a normalization constant that ensures the total area under the curve equals $1$. The beta function is also defined as: 
\begin{equation}
	B\left(\alpha, \beta\right) = \int_{0}^{1} t^{\alpha - 1}(1 - t)^{\beta - 1}dt
\end{equation}
The mean and variance of the beta distribution are calculated as follows, which perfectly control the randomly generated time steps. 
\begin{align}
	& \mu = \frac{\alpha}{\alpha + \beta} \\
	& \sigma^2 = \frac{\alpha \beta}{(\alpha + \beta)^2(\alpha + \beta + 1)}
\end{align}
\begin{figure}[!ht]
	\centering
	\includegraphics[width=0.7\textwidth]{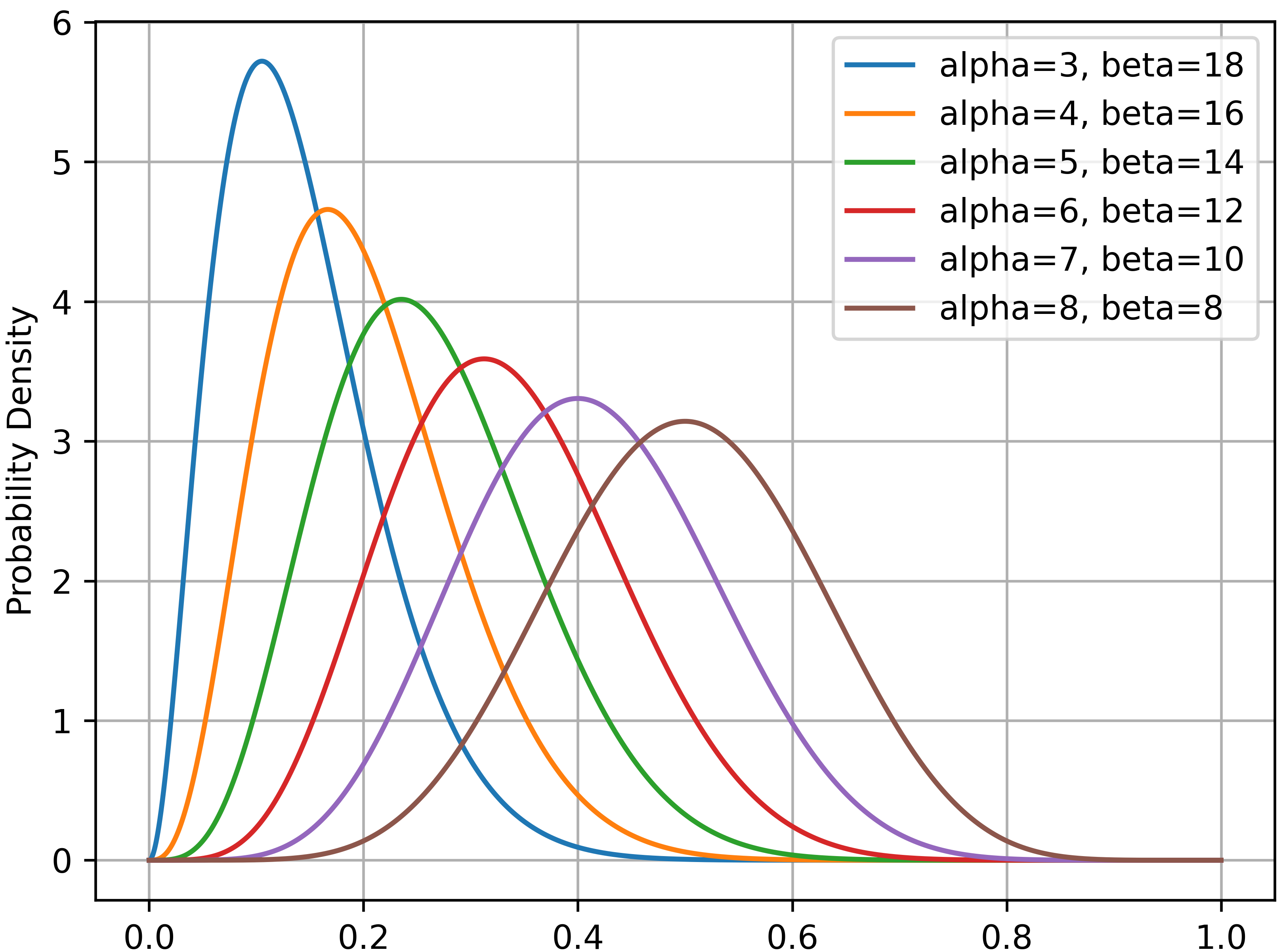}
	\caption{
        This figure displays the shapes of beta distributions generated by our scheduler for various combinations of $\alpha$ and $\beta$ values. Each curve corresponds to a distinct beta distribution, illustrating the impact of parameter variation on the distribution's shape. The curves are arranged from left to right for a model with six skip connections, progressing bottom-up from the latent space to the highest resolution of the upsampling path.
	}
	\label{fig:sub_figure04}
\end{figure}
We established a scheduler to determine the values of $\alpha$ and $\beta$, assigning distinct beta distribution functions to each level and thereby selecting diffusion time steps. This scheduling mechanism is governed by the following equations:
\begin{align}
	& \alpha = \alpha_0 + l \\
	& \beta = \beta_0 - (\beta_{c} * l)
\end{align}  
where $\alpha_{0}$, $\beta_{0}$ and $\beta_{c}$ represent initial and coefficient hyperparameters, and $l$ is the upsampling level index.  The index $l$ begins at the last skip connection and progresses towards the first skip connection, signifying the final upsampling step. Figure~\ref{fig:sub_figure04} depicts the shapes of beta distributions generated using our scheduler across varying values of $\alpha$ and $\beta$.

\section{Input Diffusion Process}\label{sec:input_diffusion_process} 
\begin{figure}[!ht]
	\centering
	\includegraphics[width=0.7\textwidth]{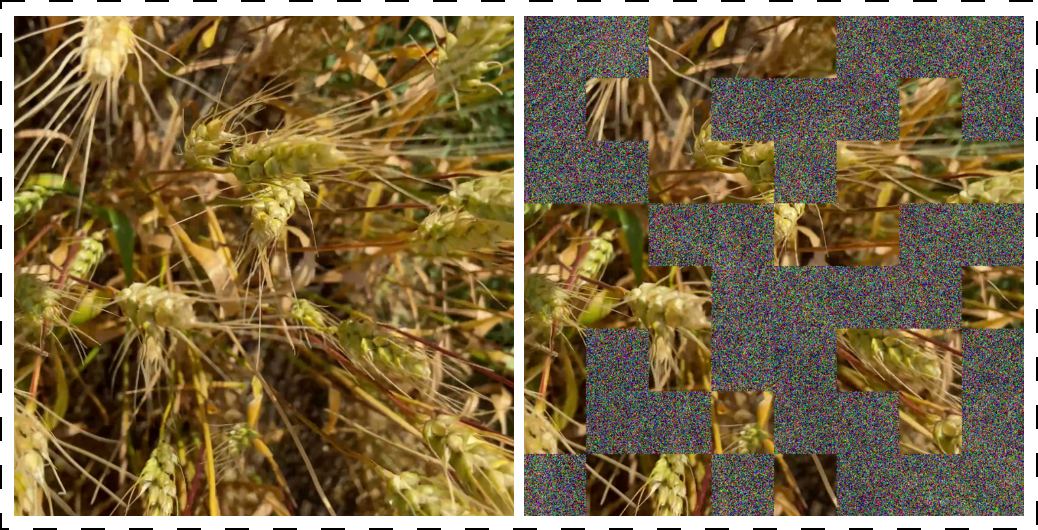}
	\caption{
		A depiction of an image frame alongside its diffused version, produced through the forward diffusion process.
	}
	\label{fig:sub_figure05}
\end{figure}

In this section, we present our strategy for diffusing input images in a patching style. Each reference frame in the mini-batch undergoes diffusion with a patching rate of $P_d = 0.5$, following the forward diffusion process with random time steps ranging from 0 to 1000. Figure~\ref{fig:sub_figure05} illustrates an example of an input image after applying this diffusion process. \par 
%

\section{Ablation Studies and Experimental Insights}\label{sec:ablation_studies} 
This section details the ablation studies performed to evaluate the performance of our proposed architecture across various configurations, both in the current setting and under alternative scenarios. Specifically, the \ImageSyntheticModel{},  \VideoSyntheticModel{}, and \XMemSyntheticModel models were developed on individual frames (only for \ImageSyntheticModel{}) and video clips, from our $\mathbb{S}_{train}$ and $\Delta$ datasets. Similarly, the \ImagePseudoModel{}, \VideoPseudoModel{}, \XMemPseudoModel were trained on individual frames (only for \ImagePseudoModel{}) and video clips from the $\PseudoLabeledVideos{train}$ and $\PseudoLabeledVideos{valid}$ datasets. All experiments are conducted under consistent configurations and settings, except for those settings that were intentionally modified, which are clearly explained in the following list: \par

\begin{enumerate}[itemsep=2ex]
    \item \textbf{QAI}: This experiment evaluates the performance of frame-level models trained for image segmentation and reconstruction tasks, where the query frame is used as input in a conventional segmentation setting. The architecture includes only the encoder, decoder, and heads, excluding the skip-spatiotemporal attention modules of the proposed video model.
    \begin{itemize}
        \item \textbf{\ImageSyntheticModel{}}: This model, evaluated without pretraining, shows moderate performance in comparison to other configurations, where a slight performance improvement is observed.
        \item \textbf{\ImagePseudoModel{}}: Pretrained on \ImageSyntheticModel{}, this configuration demonstrates a noticeable performance improvement, outperforming the previous model by a significant margin.
    \end{itemize}

    \item \textbf{PPQAI}: This experiment evaluates the model's performance when fully integrated with all components for DVOS, starting from the pretrained frame-based models. This partial pretraining allows the video models to benefit from the skip modules, which are still untrained in this configuration. Asterisks denote the best-performing models among various configurations. The full architecture is trained for the DVOS task, with the reference frame as input, and predicts the query frame and mask as output.
    \begin{itemize}
        \item \textbf{\VideoSyntheticModel{*}}: Partially pretrained with \ImageSyntheticModel{}, this model performs moderately, achieving solid results.
        \item \textbf{\VideoPseudoModel{*}}: Pretrained with \VideoSyntheticModel{*}, this model shows a significant improvement over all other models, achieving the best overall performance across all configurations.
    \end{itemize}

    \item \textbf{RRAI}: This experiment investigates the performance when frame-based models are trained with random reference frames as input. For each training sample, a single input frame is randomly selected from the set of reference frames, and the query frame and its mask are predicted in a frame-level frame/mask prediction training paradigm.
    \begin{itemize}
        \item \textbf{\ImageSyntheticModel{}}: Without pretraining, this model demonstrates relatively poor performance compared to other configurations, indicating the need for better initialization.
        \item \textbf{\ImagePseudoModel{}}: Pretrained with \ImageSyntheticModel{}, this model shows a slight improvement, though its performance still remains relatively low when compared to other, better-pretrained models.
    \end{itemize}

    \item \textbf{PPRAI}: This experiment assesses the performance when the model is partially pretrained on the RRAI \ImagePseudoModel{} model. The entire architecture is trained for the DVOS task, with the reference frame as input, and predicting the query frame and mask as output.
    \begin{itemize}
        \item \textbf{\VideoSyntheticModel{}}: Pretrained with \ImageSyntheticModel{}, this configuration provides moderate results compared to the more optimized models in other settings.
        \item \textbf{\VideoPseudoModel{}}: Pretrained with \VideoSyntheticModel{}, this model shows a noticeable performance improvement.
    \end{itemize}

    \item \textbf{FS}: This experiment evaluates DVOS models trained from scratch, without any pretraining.
    \begin{itemize}
        \item \textbf{\VideoSyntheticModel}: Trained from scratch, this model shows poor performance relative to the other models that benefit from pretraining.
        \item \textbf{\VideoPseudoModel{}}: Trained from scratch with \VideoSyntheticModel{}.
    \end{itemize}

    \item \textbf{NDNA}: This experiment tests the model's performance trained from scratch, without input and skips diffusion and without applying color augmentation.
    \begin{itemize}
        \item \textbf{\VideoSyntheticModel{}}: Without these components, the model's performance is significantly reduced, resulting in notably lower scores.
        \item \textbf{\VideoPseudoModel{}}: Despite the absence of diffusion and color augmentation, this model still shows an improvement, but its performance is relatively poorer compared to our best-performing models.
    \end{itemize}

    \item \textbf{RFI}: This experiment evaluates the model using a frame interval of 2. The first training stage on synthetic data is skipped due to the lack of natural motion in frame-per-second settings. Here, the best-performing model, \VideoSyntheticModel{*}, is used as the pretraining stage for the second stage. The model is first trained and tested with a frame interval of 2 and later tested with a frame interval of 1, identified as \textbf{RFITI1}.
    \begin{itemize}
        \item \textbf{\VideoPseudoModel{=}}: Tested with a frame interval of 2, this configuration demonstrates a slight performance drop compared to the model tested with a frame interval of 1, as seen in \textbf{RFITI1}.
    \end{itemize}

    \item \textbf{RFITI1}: This experiment tests the model with a frame interval of 1.
    \begin{itemize}
        \item \textbf{\VideoPseudoModel{=}}: With a frame interval of 1, this model shows a clear improvement compared to the performance of \textbf{RFI} tested with a frame interval of 2, suggesting better performance with a shorter frame interval.
    \end{itemize}

    \item \textbf{RRFI}: This experiment investigates random intervals of 1 or 2 for the reference frames.
    \begin{itemize}
        \item \textbf{\VideoPseudoModel{}}: Trained with reference frame intervals randomly chosen between 1 or 2, this model shows a modest performance drop compared to the best-performing model \VideoPseudoModel{*} from the PPQAI experiment.
    \end{itemize}

    \item \textbf{XMem Model}: This experiment compares our model with the XMem~\cite{cheng2022xmem} model. We used XMem as our base model due to its strong performance in VOS. XMem takes the first-frame segmentation mask as input and effectively propagates object masks across subsequent frames using a dynamic memory mechanism. Its ability to balance segmentation accuracy and computational efficiency makes it well-suited for practical applications. Furthermore, XMem remains a state-of-the-art method on multiple VOS benchmarks, providing a strong foundation for evaluating our proposed approach. As discussed in the paper, this model manipulates the given first frame's mask and generates the segmentation mask for the query mask while mainly ignoring the inputted reference and query frames, achieving higher quantitative scores by generating identical masks to the ground truth. However, in the presence of noisy input masks, it fails to predict the objects of interest, which are examined accurately through visual inspection of its predicted masks (Video 1 in Figure~\ref{fig:sub_figure06}).
    \begin{itemize}
        \item \textbf{\XMemSyntheticModel}: Evaluated with XMem-s012~\cite{cheng2022xmem} pretraining, this model shows better performance compared to other configurations of our models.
        \item \textbf{\XMemPseudoModel}: Pretrained with \XMemSyntheticModel{}, this configuration outperforms all previous configurations, setting the bar as the best-performing model.
    \end{itemize}
\end{enumerate}

\begin{table*}[!ht]
\centering
\caption{Ablation Study Results Across Different Model Configurations. This table presents the results of various ablation experiments conducted to evaluate the performance of the proposed model under different configurations. The experiments assess the impact of different model components, pretraining strategies, and training conditions on segmentation performance, measured by Dice and IoU metrics.
}
\label{table:ablation_results}
\renewcommand{\arraystretch}{1.3}
\begin{tabular}{ccccccc}
\hline
Experiment & Model & Pretrained On  & Metric & $\Psi$ & $\Gamma$ & $\PseudoLabeledVideos{test}$ \\ \hline

\multirow{4}{*}{\makecell{Query as Input \\ (QAI)}} 
& \multirow{2}{*}{\ImageSyntheticModel{}} & \multirow{2}{*}{None}                                
                        & Dice   & 0.729  & 0.408    & 0.145                        \\
                      &&& IoU    & 0.580  & 0.278    & 0.0861                       \\ \cline{3-7} 
& \multirow{2}{*}{\ImagePseudoModel{}}    & \multirow{2}{*}{\ImageSyntheticModel{}} 
                        & Dice   & 0.759  & 0.761    & 0.647                        \\
                      &&& IoU    & 0.621  & 0.619    & 0.513                        \\ \cline{2-7} 

\multirow{4}{*}{\makecell{Partially Pretrained on QAI \\ (PPQAI)}} 
& \multirow{2}{*}{\VideoSyntheticModel{*}} & \multirow{2}{*}{\ImagePseudoModel{}}    
                        & Dice   & 0.482  & 0.453    & 0.244                        \\
                      &&& IoU    & 0.335  & 0.307    & 0.150                        \\ \cline{3-7} 
& \multirow{2}{*}{\VideoPseudoModel{*}}    & \multirow{2}{*}{\VideoSyntheticModel{*}} 
                        & Dice   & \textbf{0.650}  & \textbf{0.791}    & \textbf{0.679}          \\
                      &&& IoU    & \textbf{0.493}  & \textbf{0.657}    & \textbf{0.542}          \\ \hline

\multirow{4}{*}{\makecell{Random Reference as Input \\ (RRAI)}} 
& \multirow{2}{*}{\ImageSyntheticModel{}} & \multirow{2}{*}{None}                                
                        & Dice   & 0.304  & 0.169    & 0.036                        \\
                      &&& IoU    & 0.193  & 0.100    & 0.200                        \\ \cline{3-7} 
& \multirow{2}{*}{\ImagePseudoModel{}}    & \multirow{2}{*}{\ImageSyntheticModel{}} 
                        & Dice   & 0.284  & 0.376    & 0.378                        \\
                      &&& IoU    & 0.178  & 0.248    & 0.254                        \\ \cline{2-7} 

\multirow{4}{*}{\makecell{Partially Pretrained on RAI \\ (PPRAI)}} 
& \multirow{2}{*}{\VideoSyntheticModel{}} & \multirow{2}{*}{\ImagePseudoModel{}}    
                        & Dice   & 0.487  & 0.416    & 0.315                        \\
                      &&& IoU    & 0.337  & 0.277    & 0.204                        \\ \cline{3-7} 
& \multirow{2}{*}{\VideoPseudoModel{}}    & \multirow{2}{*}{\VideoSyntheticModel{}} 
                        & Dice   & 0.611  & 0.773    & 0.672                        \\
                      &&& IoU    & 0.451  & 0.634    & 0.532                        \\ \hline

\multirow{4}{*}{\makecell{From Scratch \\ (FS)}}
& \multirow{2}{*}{\VideoSyntheticModel{}} & \multirow{2}{*}{None}    
                        & Dice   & 0.423  & 0.320    & 0.118                        \\
                      &&& IoU    & 0.290  & 0.212    & 0.069                        \\ \cline{3-7} 
& \multirow{2}{*}{\VideoPseudoModel{}}    & \multirow{2}{*}{\VideoSyntheticModel{}} 
                        & Dice   & 0.647  & 0.782    & 0.670                        \\
                      &&& IoU    & 0.489  & 0.647    & 0.535                        \\ \hline

\multirow{4}{*}{\makecell{No Diffusion \& No Color Augmentation \\ (NDNA)}}
& \multirow{2}{*}{\VideoSyntheticModel{}} & \multirow{2}{*}{None}    
                        & Dice   & 0.347  & 0.080    & 0.018                        \\
                      &&& IoU    & 0.232  & 0.046    & 0.010                        \\ \cline{3-7} 
& \multirow{2}{*}{\VideoPseudoModel{}}    & \multirow{2}{*}{\VideoSyntheticModel{}} 
                        & Dice   & 0.664  & 0.578    & 0.675                        \\
                      &&& IoU    & 0.504  & 0.429    & 0.541                        \\ \hline

\multirow{2}{*}{\makecell{Reference Frame Interval of $2$ \\ (RFI)}}
& \multirow{2}{*}{\VideoPseudoModel{=}}    & \multirow{2}{*}{\VideoSyntheticModel{*}} 
                        & Dice   & 0.555  & 0.735    & 0.622                        \\
                      &&& IoU    & 0.398  & 0.586    & 0.476                        \\ \cline{3-7} 

\multirow{2}{*}{\makecell{RFI Tested on Interval of 1 (RFITI1)}}
& \multirow{2}{*}{\VideoPseudoModel{=}}    & \multirow{2}{*}{\VideoSyntheticModel{*}} 
                        & Dice   & 0.645  & 0.780    & 0.671                        \\
                      &&& IoU    & 0.484  & 0.641    & 0.532                        \\ \hline

\multirow{2}{*}{\makecell{Random Reference Frame Intervals of $1$ or $2$ \\ (RRFI)}}
& \multirow{2}{*}{\VideoPseudoModel{}}    & \multirow{2}{*}{\VideoSyntheticModel{*}} 
                        & Dice   & 0.533  & 0.741    & 0.623                        \\
                      &&& IoU    & 0.383  & 0.595    & 0.478                        \\ \hline\hline

\multirow{4}{*}{\makecell{XMem Model~\cite{cheng2022xmem}}}
& \multirow{2}{*}{\XMemSyntheticModel} & \multirow{2}{*}{XMem-s012~\cite{cheng2022xmem}}    
                        & Dice   & 0.794 & 0.454 & 0.448                        \\
                      &&& IoU    & 0.668 & 0.314 & 0.302                       \\ \cline{3-7} 
& \multirow{2}{*}{\XMemPseudoModel}    & \multirow{2}{*}{\XMemSyntheticModel} 
                        & Dice   & \textbf{0.831}  & \textbf{0.811} & \textbf{0.835}      \\
                      &&& IoU    & \textbf{0.716}  & \textbf{0.690} & \textbf{0.726}       \\ \hline

\end{tabular}
\end{table*}

\begin{figure*}[!ht]
	\centering
	\includegraphics[width=\textwidth]{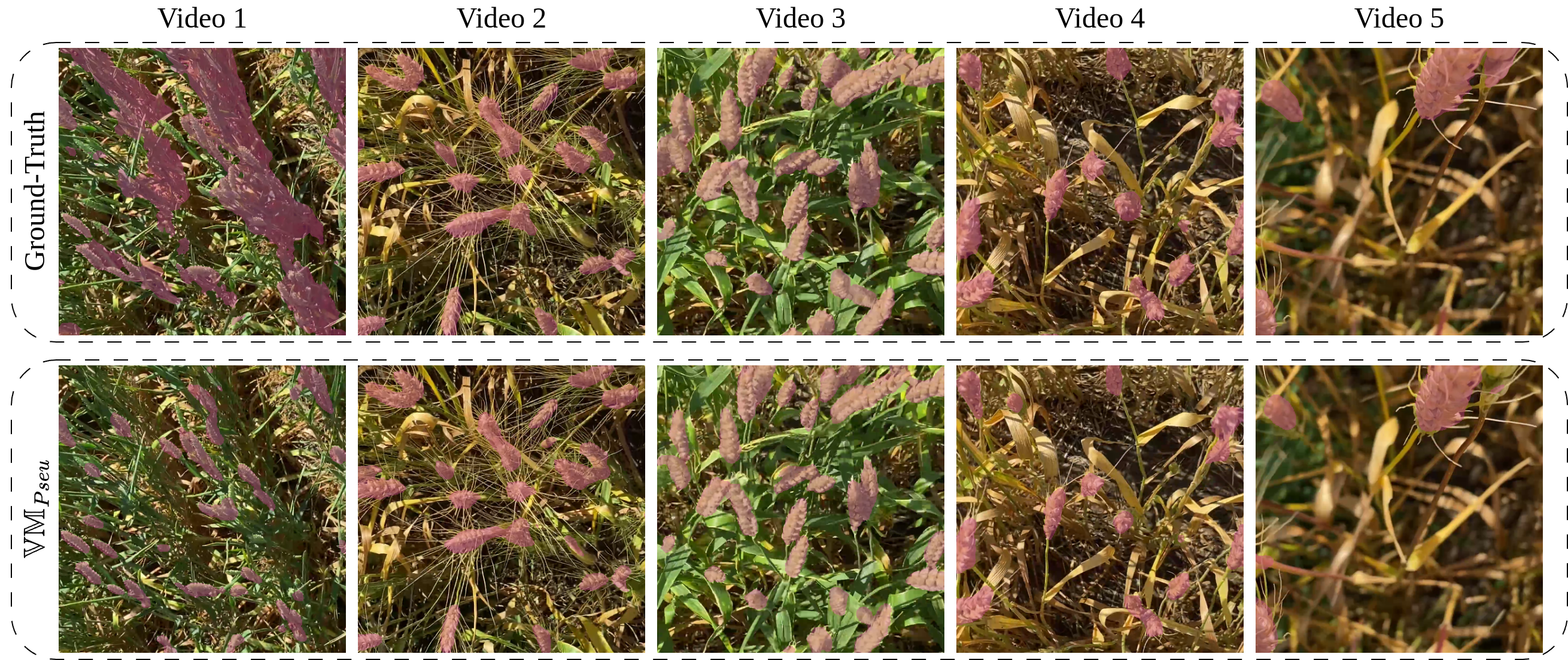}
	\caption{Segmentation prediction performance of \VideoPseudoModel{*} across various videos in $\PseudoLabeledVideos{test}$. 
	}
	\label{fig:sub_figure06}
\end{figure*}

\end{document}